\documentclass[11pt]{article}
\usepackage[preprint]{acl}

\usepackage{times}
\usepackage{latexsym}
\usepackage{amsfonts}
\usepackage{amsmath}
\usepackage{amssymb}
\usepackage{multirow}
\usepackage{booktabs}
\usepackage[table]{xcolor}
\usepackage{enumitem}
\definecolor{lightpurple}{RGB}{238,235,255}
\definecolor{deepred}{RGB}{180,0,0}

\usepackage[T1]{fontenc}
\usepackage[utf8]{inputenc}

\usepackage{microtype}

\usepackage{inconsolata}

\usepackage{graphicx}
\usepackage{booktabs}
\usepackage{graphicx}
\usepackage[most]{tcolorbox}
\usepackage{xcolor}
\usepackage{caption}
\usepackage{algorithm}
\usepackage{algpseudocode}

\definecolor{yardorange}{RGB}{225,165,35}
\definecolor{yardblue}{RGB}{40,70,220}
\definecolor{framegold}{RGB}{230,210,110}
\definecolor{hallred}{RGB}{220,45,45}
\definecolor{lightbg}{RGB}{255,254,248}

\newcommand{\hall}[1]{\textcolor{hallred}{#1}}

\newtcolorbox{casebox}{
  enhanced,
  colback=lightbg,
  colframe=framegold,
  boxrule=0.45pt,
  arc=1pt,
  left=3pt,
  right=3pt,
  top=2pt,
  bottom=2pt,
  boxsep=1pt,
  width=\linewidth,
  before skip=0pt,
  after skip=0pt
}

\title{YARD: Y-Architecture Register Decoding for Efficient Hallucination Mitigation in Large Vision-Language Models}

\author{
\normalfont
\textbf{Ting Chen\textsuperscript{1,*}} \quad
\textbf{Geng Li\textsuperscript{2,*}} \quad
\textbf{Guohao Chen\textsuperscript{2,*}} \quad
\textbf{Yu Hu\textsuperscript{1,\ensuremath{\dagger}}}
\\
\textbf{Guan Huang\textsuperscript{3}} \quad
\textbf{Mai Chen\textsuperscript{3}} \quad
\textbf{Langsheng Lei\textsuperscript{3}} \quad
\textbf{Jun Du\textsuperscript{3}}
\\
\textsuperscript{1}Guangdong University of Technology \quad
\textsuperscript{2}Nanyang Technological University
\\
\textsuperscript{3}Shenzhen TENCLASS Technology Co., Ltd.
\\
\small{\textsuperscript{*}Equal contribution. \textsuperscript{\ensuremath{\dagger}}Corresponding author.}
}

\begin{document}
\maketitle
\begin{abstract}

Contrastive decoding (CD) seeks to mitigate hallucinations in Large Vision-Language Models (LVLMs) by contrasting the output distributions of a standard model and a visually degraded model. However, existing training-free CD methods suffer from sub-optimal degraded branches: completely dropping visual tokens is too extreme and induces language hallucinations, while corrupting input images offers coarse control over visual evidence and suffers from high inference latency due to requiring two full forward passes. To address these dilemmas, we propose \textbf{YARD}, a training-free \textbf{Y-Architecture Register Decoding} framework. Motivated by the observation that reliable text-to-vision grounding predominantly emerges in the middle decoder layers, YARD constructs the degraded branch internally by sharing shallow-layer computations and branching exactly at this critical stage. For the degraded branch, YARD replaces patch-level visual tokens with register tokens, which preserve global image semantics but lack fine-grained local evidence. This image-aware yet locally under-grounded design provides a faithful contrastive signal without extreme modality mismatch, while the Y-architecture strictly avoids a costly second forward pass. Extensive experiments on generative and discriminative hallucination benchmarks demonstrate that YARD consistently achieves state-of-the-art hallucination mitigation across multiple LVLMs, alongside a significant reduction in inference latency.

% Large vision-language models (LVLMs) often generate fluent but visually unsupported responses, making hallucination mitigation essential for reliable multimodal generation. Most of existing training-free contrastive decoding methods usually construct degraded branches by removing visual information or corrupting input images, which are either image-agnostic or too coarse to isolate hallucination-prone signals. We propose \textbf{YARD}, a training-free \textbf{Y-Architecture Register Decoding} framework that constructs the degraded branch inside the LLM decoder. Based on the observation that reliable text-to-vision grounding emerges in the middle decoder layers, YARD shares shallow-layer computation and branches at this critical stage. The clean branch retains patch-level visual tokens, while the degraded branch replaces them with register representations that preserve global image semantics but lack fine-grained local grounding. This image-aware yet locally under-grounded branch provides a more faithful contrastive signal while avoiding a full second forward pass. Experiments on generative and discriminative hallucination benchmarks show that YARD consistently reduces hallucinations across multiple LVLM architectures with plug-and-play efficiency.
\end{abstract}

\section{Introduction}

Large vision-language models (LVLMs)~\cite{liu2024improved,dai2023instructblip,wang2024qwen2,zhu2024minigpt} connect visual encoders with large language models, enabling strong visual understanding and language generation capabilities across visual question answering, image captioning, and open-ended multimodal dialogue. However, despite their ability to produce fluent and contextually coherent responses, hallucination remains a central obstacle to their reliability\cite{li2023evaluating}. LVLMs may generate text that is semantically plausible but inconsistent with the visual content, such as non-existent objects\cite{rohrbach2018object}, incorrect attributes\cite{pham2024h}, or spatial relations unsupported by the image\cite{guan2024hallusionbench}. Such failures suggest that generation is not always faithfully grounded in the visual evidence, making hallucination mitigation a key challenge for reliable LVLM deployment\cite{bai2024hallucination}.

\begin{figure}[t]
\centering
\includegraphics[width=\columnwidth]{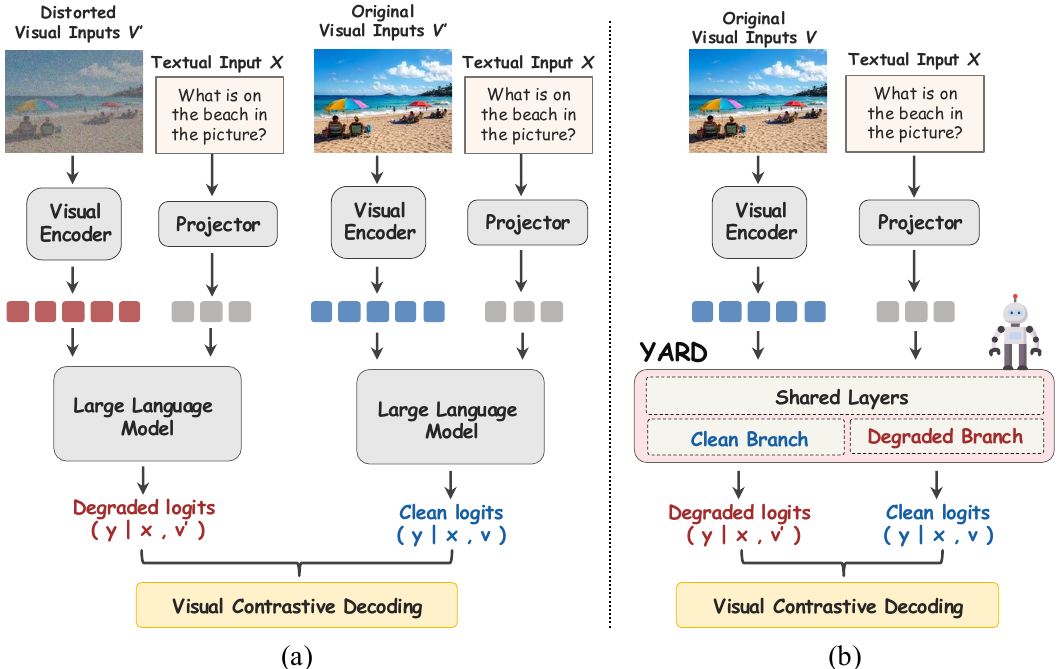}
\caption{Comparison between (a) pixel-level degradation that corrupts the visual input and (b) YARD that constructs an image-aware but locally under-grounded degraded branch inside the LLM decoder.}
\label{fig:teaser}
\vspace{-15pt}
\end{figure}
Among existing training-free approaches to hallucination mitigation, contrastive decoding~\cite{chuang2024dola} can be viewed as a form of hallucination-component separation at the logit level.
Specifically, the clean logits of an LVLM contain both predictions supported by genuine visual evidence and hallucination-related components induced by language priors or insufficient grounding.
The role of the degraded branch is to construct a contrastive distribution that exposes these hallucination-prone components, so that they can be selectively suppressed through logit subtraction~\cite{li2023contrastive}.

Existing methods typically instantiate the degraded branch in two ways.
\textit{One line of work masks or removes visual information inside the LLM,} reducing the degraded branch to an approximately text-only language-prior branch~\cite{favero2024multi,zhu2025ibd,wang2024mitigating,woo2025don,huo2024self,deng2025maskcd,manevich2024mitigating,li2025mitigating}.
However, once visual conditioning is completely removed, the induced hallucinations often arise from generic language priors and may not align with the input image.
\textit{Another line perturbs the original image in pixel space by adding noise, masking image regions,} or corrupting visual details~\cite{chen2024halc,yang2024pensieve,zhao2025cross}.
However, pixel-level degradation cannot selectively disrupt visual signals: color, texture, object boundaries, spatial layout, and local semantics are tightly entangled in pixel space.
As a result, such perturbations may fail to isolate the local evidence responsible for hallucination and instead contaminate the degraded logits with signals associated with genuine visual evidence.

These limitations motivate us to move degradation into a more structured feature-level space, where global image semantics can be preserved while fine-grained local visual evidence is selectively weakened.
Table~\ref{tab:degradation_alpha} further supports this choice, showing that feature-level degradation exposes a purer hallucination signal than text-only or pixel-level degradation.
However, existing contrastive decoding methods mainly construct degraded branches from either the language side or the visual-input side, leaving feature-level degradation inside the LVLM decoder largely underexplored.
This raises two key design questions: \textbf{\textit{where}} should the degraded branch be constructed inside the LVLM, and \textbf{\textit{how}} should visual evidence be degraded at the feature level?
\begin{figure*}[t]
    \centering
    \includegraphics[width=0.98\textwidth]{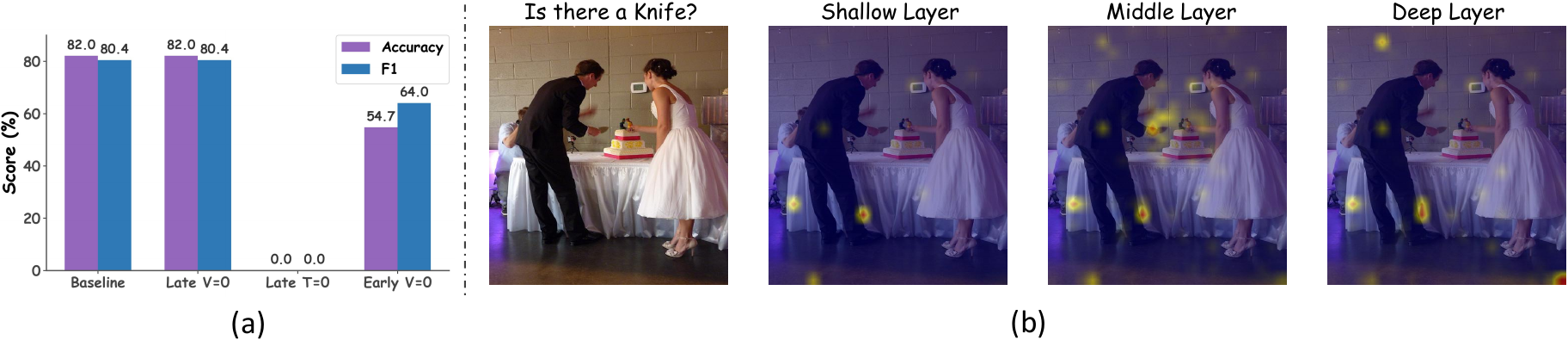}
\caption{
Motivating analysis of visual evidence flow.
(a) Zero-out intervention shows that late-layer predictions mainly rely on text-side representations.
(b) Text-to-vision attention becomes more grounded at the middle layer, e.g., layer 10, indicating the key window for local visual evidence extraction.
}
    \label{fig:motivation}
\vspace{-10pt}
\end{figure*}

To answer this question, we analyze the cross-modal information flow between visual and text tokens in the decoder\cite{zhang2025cross,jiang2025devils}.
We observe that perturbing visual tokens near the output layers has only a limited impact on final predictions, whereas perturbing text tokens leads to severe generation collapse\cite{chen2024image,fan2025visipruner}.
This suggests that, in deeper layers, task-relevant visual evidence has largely been integrated into text-side representations.
Meanwhile, the layer-wise evolution of text-to-vision attention shows that reliable visual grounding mainly emerges in the middle layers.
Therefore, as illustrated in Figure~\ref{fig:motivation}, the degraded branch should be constructed around the \textbf{middle decoder layers}.
Furthermore, Table~\ref{tab:degradation_alpha} shows that register-based degradation provides the most effective degraded branch.
This leads to the second design question: \textbf{\textit{how should we degrade the visual condition after branching?}}
Table~\ref{tab:degradation_alpha} shows that \textbf{Global Info}. achieves better hallucination mitigation than Local Info., which still relies on residual local visual evidence.
This suggests that a more suitable degraded branch should preserve the global semantics of the input image, so that \textbf{\textit{it remains image-relevant, while removing fine-grained local evidence to prevent reliable local grounding.}}
Such a global-but-local-missing condition can more effectively expose hallucination-prone signals.

Based on these analyses, we propose \textbf{YARD}, a training-free Y-Architecture contrastive decoding framework.
YARD realizes \textbf{feature-level} degradation from two perspectives: \textbf{where} to degrade and \textbf{how} to degrade.
First, for where to degrade, \textbf{YARD} shares the shallow decoder layers between the clean and degraded branches and splits them at a middle layer.
This allows the degraded branch to inherit early multimodal context while preventing it from fully integrating local visual evidence.
Second, for \textbf{how} to degrade, YARD constructs a feature-level visual condition that preserves \textbf{global information} while removing \textbf{local information}.
Specifically, prior studies~\citep{darcet2024vision} have observed abnormally high-norm \textit{attention-sink} tokens in vision encoders, which tend to absorb and carry coarse image-level information.
YARD applies sink-shift to redirect these sink activations to newly introduced register tokens, enabling them to inherit the role of global-information carriers.
These register representations are then used to replace fine-grained visual tokens in the degraded branch.
In addition, \textbf{YARD} improves inference efficiency by avoiding a complete second forward pass.
Empirically, \textbf{YARD} consistently reduces hallucinations in both generative and discriminative evaluations.
These gains transfer across diverse LVLM architectures, demonstrating that \textbf{YARD} provides a general training-free contrastive signal rather than an architecture-specific decoding heuristic.
Our contributions are summarized as follows:
\begin{itemize}[leftmargin=*]
    \item We propose register-based feature-level degradation to construct an image-aware but locally under-grounded contrastive branch without corrupting the visual input.

    \item We analyze degraded-branch purity and cross-modal information flow, motivating faithful degradation at middle decoder layers.

    \item We present \textbf{YARD}, an efficient training-free Y-Architecture contrastive decoding framework that consistently mitigates hallucinations across benchmarks and LVLMs.
\end{itemize}

\section{Related Work}

\subsection{Contrastive Decoding for Hallucination Mitigation}

Training-free contrastive decoding mitigates hallucinations by contrasting a clean branch with a hallucination-prone degraded branch, and suppressing predictions overly favored by the degraded condition. Existing methods mainly differ in how this degraded branch is constructed.

One line of work applies degradation at the input level. VCD~\cite{leng2024mitigating} perturbs images with Gaussian noise, while ICD~\cite{wang2024mitigating}, DCD~\cite{wu2025mitigating}, VACoDe~\cite{kim2024vacode}, and Octopus\cite{suo2025octopus} introduce instruction, multi-source, or adaptive perturbations. These methods are effective but typically require largely separate clean and degraded computation paths, and their degradation is only indirectly related to the internal grounding process.

Another line of work obtains contrastive or corrective signals from internal states or decoding dynamics. DoLa~\cite{chuang2024dola} contrasts logits across decoder layers, LayerCD~\cite{tong2025mitigating} extends layer-wise contrast to visual representations, and OPERA~\cite{huang2024opera} and ECD~\cite{fieback2025efficient} use decoding-time regularization or auxiliary scoring. While avoiding explicit input corruption, they do not explicitly construct a degraded visual branch at the stage where text-to-vision grounding is formed.

In contrast, YARD constructs the degraded branch inside the LLM decoder. By branching at the middle layers, it aligns degradation with the formation of visual grounding while sharing shallow-layer computation between clean and degraded branches.

\subsection{Register Tokens in Vision Transformers}

Register tokens were introduced to address high-norm outlier artifacts in Vision Transformers\cite{xiao2024efficient,sun2024massive}. 
Darcet et al.~\cite{darcet2024vision} showed that such outliers often emerge in spatially low-informative regions of ViTs such as DINOv2\cite{oquab2023dinov2} and CLIP\cite{radford2021learning}, and proposed learnable registers to absorb them. 
Jiang et al.~\cite{jiang2026vision} further showed that register-like tokens can be constructed at test time without retraining by redirecting anomalous MLP activations to an appended token.

Different from prior work that uses registers to stabilize clean visual representations, YARD uses them as degraded visual conditions for contrastive decoding. 
Since register representations preserve coarse global semantics but lack stable patch-level correspondence, they naturally form an image-aware but locally unreliable branch for hallucination mitigation.

\section{Preliminaries}
\label{sec:prelim}

\paragraph{Training-free register tokens.}
Given an image, a Vision Transformer produces patch tokens
$\mathbf{Z}=\{z_1,\ldots,z_N\}$, where each $z_j\in\mathbb{R}^{d}$ represents a local image patch. Prior work shows that ViTs can produce high-norm outlier tokens at spatially uninformative regions, and introduces register tokens to absorb such artifacts~\cite{darcet2024vision}. Jiang et al.~\cite{jiang2026vision} further show that register tokens can be constructed at test time without retraining, by appending an extra token and redirecting anomalous MLP activations to it.

In this work, we use register tokens not to improve the clean visual representation, but to construct degraded visual conditions for contrastive decoding. Compared with ordinary patch tokens, register tokens preserve coarse global visual semantics but lack stable patch-level correspondence and fine-grained local details. This makes them suitable for forming a branch that remains image-aware while being locally under-grounded. Details are provided in Appendix~\ref{app:register_implementation}.

\begin{figure*}[t]
    \centering
    \includegraphics[width=0.98\textwidth]{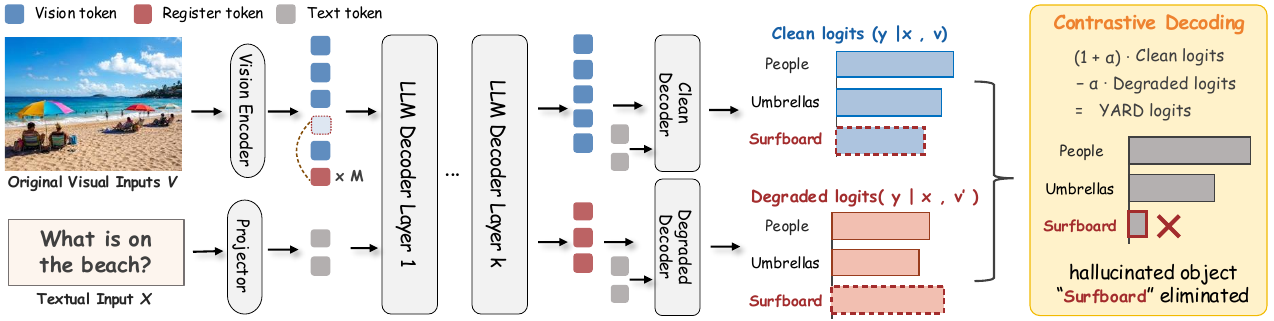}
\caption{
Overview of \textbf{Y-A}rchitecture \textbf{R}egister \textbf{D}ecoding.
}
    \label{fig:method}
\end{figure*}

\section{Motivating Analysis}
\paragraph{Observation 1: Middle Layers Are the Critical Window for Visual Evidence Transfer}
\label{sec:where_to_degrade}

We first examine where a degraded branch should be constructed inside the LVLM decoder.
Although vision tokens provide the initial visual source, final generation may increasingly rely on text-token representations, as text tokens progressively absorb task-relevant visual evidence through causal self-attention.

To verify this, we zero out either vision or text hidden states at decoder layer $\ell$:
\begingroup
\small
\setlength{\abovedisplayskip}{3pt}
\setlength{\belowdisplayskip}{3pt}
\setlength{\abovedisplayshortskip}{2pt}
\setlength{\belowdisplayshortskip}{2pt}
\begin{equation}
\mathbf{H}^{(\ell)}_{\setminus Z}=[\mathbf{0};\mathbf{T}^{(\ell)}],
\quad
\mathbf{H}^{(\ell)}_{\setminus T}=[\mathbf{Z}^{(\ell)};\mathbf{0}],
\end{equation}
\endgroup
where {\small $\mathbf{H}^{(\ell)}=[\mathbf{Z}^{(\ell)};\mathbf{T}^{(\ell)}]$}.
The intervened states are then passed through the remaining decoder layers.

As shown in Figure~\ref{fig:motivation}(a), zeroing out vision tokens at the first layer leads to a clear performance drop, confirming that vision tokens are indispensable as the source of visual evidence.
However, zeroing out vision tokens near the output layers has only a limited effect on final predictions.
In contrast, \textbf{\textit{zeroing out text tokens causes complete generation failure}}, with both accuracy and F1 dropping to \textbf{0}.
This suggests that, by the late decoder layers, task-relevant visual evidence has largely been transferred into text-side representations.
Therefore, perturbing vision tokens after this transfer provides only a limited basis for constructing an effective degraded branch.

This finding further raises a layer-wise question: when does visual evidence become effectively transferred into text tokens?
To answer this, we visualize text-to-vision attention across decoder layers by projecting it back to the image space.
As shown in Figure~\ref{fig:motivation}(b), shallow-layer attention is diffuse and often distracted by irrelevant regions, whereas middle-layer attention becomes concentrated on task-relevant objects.
This suggests that the middle layers form the key window where text tokens begin to acquire local visual evidence before it is fully internalized in deeper text-side representations.
Therefore, the degraded branch should be constructed around the middle decoder layers: branching too early lacks sufficient cross-modal context, while branching too late leaves limited room for weakening visual evidence.
This analysis determines \textbf{\emph{where}} to degrade; we next examine \textbf{\emph{how}} to construct the degraded branch to better expose hallucination-related signals.

\begin{table}[h]
\centering
\small
\setlength{\tabcolsep}{3.0pt}
\renewcommand{\arraystretch}{1.08}
\caption{
Ablation of degraded-branch construction.
Pixel-level corrupts the input image; Text-only removes visual tokens; Local Info. randomly keeps a subset of vision tokens; Global Info. replaces fine-grained visual tokens with register representations.
Parentheses report the hallucination increase after removing $\alpha$. $\mathrm{CH}_{s}$ and $\mathrm{CH}_{i}$ are CHAIR-based metrics for evaluating hallucinations in image captions.
}
\label{tab:degradation_alpha}
\begin{tabular}{lcccc}
\toprule
\multirow{2}{*}{Degradation}
& \multicolumn{2}{c}{w/ $\alpha$}
& \multicolumn{2}{c}{\cellcolor{lightpurple}w/o $\alpha$} \\
\cmidrule(lr){2-3} \cmidrule(lr){4-5}
& CH$_s\downarrow$ & CH$_i\downarrow$
& \cellcolor{lightpurple}CH$_s\downarrow$
& \cellcolor{lightpurple}CH$_i\downarrow$ \\
\midrule
Pixel-level
& 23.0 & 8.3
& \cellcolor{lightpurple}28.7 {\color{deepred}\textbf{(+5.7)}}
& \cellcolor{lightpurple}12.1 {\color{deepred}\textbf{(+3.8)}} \\

Text-only
& \underline{21.7} & \underline{7.0}
& \cellcolor{lightpurple}\underline{24.0} {\color{deepred}\textbf{(+2.3)}}
& \cellcolor{lightpurple}\underline{7.9} {\color{deepred}\textbf{(+0.9)}} \\

\addlinespace[1pt]
\rowcolor{gray!15}
\multicolumn{5}{c}{\textit{Feature-level degraded conditions}} \\
\addlinespace[1pt]

Local Info.
& \underline{25.0} & \underline{7.9}
& \cellcolor{lightpurple}\underline{26.4} {\color{deepred}\textbf{(+1.4)}}
& \cellcolor{lightpurple}\underline{8.4} {\color{deepred}\textbf{(+0.5)}} \\

Global Info.
& \textbf{19.7} & \textbf{6.5}
& \cellcolor{lightpurple}\textbf{21.0} {\color{deepred}\textbf{(+1.3)}}
& \cellcolor{lightpurple}\textbf{7.0} {\color{deepred}\textbf{(+0.5)}} \\
\bottomrule
\end{tabular}
\end{table}

\paragraph{Observation 2: Feature-level degradation exposes a purer hallucination signal.}
Table~\ref{tab:degradation_alpha} compares different degraded-branch constructions from the perspective of signal purity.
To more strictly examine whether the degraded branch precisely exposes hallucination-prone logits, we remove the conservative clean-logit anchoring controlled by $\alpha$ in standard contrastive decoding and adopt a direct subtraction form:
\vspace{-5pt}
\begin{equation}
\ell_i^{\mathrm{w/o}\ \alpha} = \ell_i^c - \ell_i^d .
\end{equation}
\vspace{-5pt}
\textbf{\textit{This setting is not intended to achieve the best decoding performance, but instead serves as a stress test for the degraded branch.}}
Ideally, the clean logits contain both correct predictions supported by genuine visual evidence and hallucination-prone predictions caused by insufficient grounding, while the degraded logits should be concentrated on the latter.
In this case, even without the additional clean-branch protection introduced by $\alpha$, direct subtraction should mainly suppress hallucination-related components without substantially damaging visually grounded predictions.

The results show that \textbf{pixel-level degradation} produces the largest increase in CH$_s$/CH$_i$ under the w/o $\alpha$ setting, indicating that \textbf{\textit{its degraded branch is contaminated by residual visual semantics and fails to precisely isolate hallucination components.}}
\textbf{Text-only degradation} leads to a smaller increase, but this mainly comes from its image-agnostic nature after completely removing visual conditioning.
Since its degraded logits are weakly aligned with the clean prediction for the current image,\textbf{\textit{ they contain fewer image-specific correct object signals as well as fewer image-specific hallucination signals; }}therefore, direct subtraction is less likely to damage visually grounded predictions, but the resulting contrastive signal is also \textbf{less aligned with} the hallucination components in the clean branch.

In contrast, \textbf{feature-level degradation} exhibits the smallest drop after removing $\alpha$, suggesting that its register-conditioned branch induces the purest hallucination signal.
By preserving global image semantics while weakening local grounding, its degraded logits contain fewer correct object signals supported by genuine visual evidence and are more \textbf{\textit{concentrated on the hallucination component targeted by contrastive subtraction.}}
Among feature-level variants, Table~\ref{tab:degradation_alpha} shows that Global Info. outperforms Local Info. as the degraded condition.
Local Info., obtained by randomly retaining a subset of vision tokens, may leak residual patch-level evidence into the degraded branch.
In contrast, Global Info. preserves coarse image-level semantics while removing fine-grained local evidence, forming a cleaner global-but-local-missing condition.
This explains why it achieves the lowest CH$_s$/CH$_i$ and the smallest hallucination increase after removing $\alpha$.

\section{Method}
\label{sec:yard}

The observations above motivate a simple principle: rather than degrading the visual source itself, the degraded branch should intervene in the process by which text tokens extract fine-grained visual evidence from vision tokens.
To this end, we propose \textbf{YARD}, a training-free Y-Architecture contrastive decoding framework that constructs the degraded branch inside the LLM decoder.
The clean and degraded branches share the shallow decoder layers and split at a middle layer, where text-to-vision grounding has begun to emerge but fine-grained local evidence has not yet been fully integrated.
After branching, the clean branch retains the original patch-level visual tokens, while the degraded branch discards them and conditions generation on register representations together with text tokens.
This produces a branch that remains globally image-aware but locally under-grounded.

Given an image, the vision encoder produces patch-level visual tokens
$\mathbf{Z}=\{z_1,\ldots,z_N\}$.
We construct training-free register tokens
$\mathbf{R}=\{r_1,\ldots,r_M\}$ following Section~\ref{sec:prelim}, and denote the text tokens as
$\mathbf{T}=\{t_1,\ldots,t_L\}$.
The input sequence to the LLM decoder is
\begin{equation}
\mathbf{X}^{(0)}
=
[\mathbf{Z};\mathbf{R};\mathbf{T}].
\end{equation}
Here, $\mathbf{Z}$ provides fine-grained local visual evidence, while $\mathbf{R}$ provides a coarse but spatially imprecise visual condition.

Let the LLM decoder contain $D$ layers, and let $K$ denote the branching layer.
YARD first processes the full sequence through the shared prefix:
\begin{equation}
\mathbf{H}^{(K)}
=
F_{1:K}(\mathbf{X}^{(0)})
=
[\mathbf{Z}^{(K)};\mathbf{R}^{(K)};\mathbf{T}^{(K)}].
\end{equation}
This shared computation allows both branches to inherit the same early multimodal context, while avoiding the full second forward pass required by input-level contrastive decoding.

At layer $K$, YARD splits the computation into a clean branch and a degraded branch:
\begin{equation}
\begin{aligned}
\mathbf{H}^{(K)}_{c}
&= [\mathbf{Z}^{(K)};\mathbf{R}^{(K)};\mathbf{T}^{(K)}], \\
\mathbf{H}^{(K)}_{d}
&= [\mathbf{R}^{(K)};\mathbf{T}^{(K)}].
\end{aligned}
\end{equation}
The clean branch keeps the original visual tokens and therefore preserves access to fine-grained patch-level evidence.
The degraded branch removes these patch tokens and relies on register representations, which preserve coarse image-level semantics but lack stable local correspondence.

The two branches are then forwarded through the remaining decoder layers:
\begin{equation}
\begin{aligned}
\mathbf{H}^{(D)}_{c}
&= F_{K+1:D}(\mathbf{H}^{(K)}_{c}), \\
\mathbf{H}^{(D)}_{d}
&= F_{K+1:D}(\mathbf{H}^{(K)}_{d}).
\end{aligned}
\end{equation}
Let $\mathbf{h}^{(D)}_{c,i}$ and $\mathbf{h}^{(D)}_{d,i}$ denote the hidden states used to predict the next token at step $i$ in the clean and degraded branches, respectively.
The vocabulary-level logits are obtained through the language modeling head $g_{\mathrm{lm}}$:
\begin{equation}
\boldsymbol{\ell}^{c}_{i}
=
g_{\mathrm{lm}}(\mathbf{h}^{(D)}_{c,i}),
\qquad
\boldsymbol{\ell}^{d}_{i}
=
g_{\mathrm{lm}}(\mathbf{h}^{(D)}_{d,i}).
\end{equation}
YARD then applies contrastive decoding with the register-conditioned degraded logits:
\begin{equation}
\boldsymbol{\ell}^{\mathrm{yard}}_{i}
=
(1+\alpha)\boldsymbol{\ell}^{c}_{i}
-
\alpha\boldsymbol{\ell}^{d}_{i},
\end{equation}
where $\alpha$ controls the contrastive strength.
The final decoding distribution is obtained by applying softmax to $\boldsymbol{\ell}^{\mathrm{yard}}_{i}$.
Following standard contrastive decoding practice, we use a clean-branch plausibility constraint to avoid over-promoting tokens with very low clean probability.

This design creates a dual degradation effect.
First, the degraded branch inherits the shared shallow-to-middle representations, where text-to-vision grounding has begun but has not yet fully integrated fine-grained local evidence.
Second, after branching, it is explicitly deprived of patch-level visual tokens and can only rely on register-level global semantics.
As a result, the degraded branch remains semantically coherent with the image, but lacks the local visual evidence required for faithful grounding.
Predictions that are plausible under this locally under-grounded branch, but insufficiently supported by the clean branch, are therefore suppressed during contrastive decoding.

\begin{algorithm}[t]
\captionsetup{font=small}
\caption{YARD Contrastive Decoding}
\label{alg:yard}
\small
\begin{algorithmic}[1]
\Require Image $I$, text query $T$, branching layer $K$, contrastive strength $\alpha$
\Ensure Model response with register-conditioned contrastive decoding
\State Encode $I$ into patch tokens $\mathbf{Z}$ and register tokens $\mathbf{R}$
\State Run shared decoder prefix: $\mathbf{H}^{(K)} = F_{1:K}([\mathbf{Z};\mathbf{R};\mathbf{T}])$
\State Split branches: $\mathbf{H}^{(K)}_c=[\mathbf{Z}^{(K)};\mathbf{R}^{(K)};\mathbf{T}^{(K)}]$, $\mathbf{H}^{(K)}_d=[\mathbf{R}^{(K)};\mathbf{T}^{(K)}]$
\State Obtain $\boldsymbol{\ell}^c_i,\boldsymbol{\ell}^d_i$ from remaining layers and decode with $\boldsymbol{\ell}^{\mathrm{yard}}_i=(1+\alpha)\boldsymbol{\ell}^c_i-\alpha\boldsymbol{\ell}^d_i$
\end{algorithmic}
\end{algorithm}

\section{Experiments}

We evaluate YARD on both generative and discriminative hallucination benchmarks. 
For generative evaluation, we use AMBER\cite{wang2023amber}, Object HalBench\cite{yu2024rlhf}, and MME-Hallucination\cite{fu2026mme}; for discriminative evaluation, we use AMBER discrimination\cite{wang2023amber} and POPE\cite{li2023evaluating}. 
Using LLaVA-1.5\cite{liu2024improved} as the main backbone, we compare YARD with representative training-free methods, including ICD\cite{wang2024mitigating}, OPERA\cite{huang2024opera}, VCD\cite{leng2024mitigating}, M3ID\cite{favero2024multi}, AVISC\cite{woo2025don}, EVAS~\citep{zhang2025shallow}, FuzzyCD\cite{kim2025fuzzy}, and TAME~\citep{tang2025intervening}. 
We further evaluate transferability on LLaVA-NeXT\cite{liu2024llavanext}, Qwen2-VL\cite{wang2024qwen2}, Qwen3-VL\cite{bai2025qwen3}, InstructBLIP\cite{dai2023instructblip}, and Mini-Gemini\cite{li2025mini}.

\begin{table*}[t]
\centering
\caption{Generative hallucination evaluation on LLaVA-1.5-7B across AMBER, Object HalBench, and MME. Cov. measures object coverage, Hal measures the response-level hallucination rate, Cog reflects the tendency to generate plausible but image-unfaithful objects; 
Exist. and Pos. denote MME sub-scores for existence and position.
}
\vspace{-3pt}
\label{tab:hallucination_all}
\scriptsize
\setlength{\tabcolsep}{3.5pt}
\renewcommand{\arraystretch}{1.0}
\resizebox{\textwidth}{!}{
\begin{tabular}{lcccc cc ccccc}
\toprule
\multirow{2}{*}{Method}
& \multicolumn{4}{c}{AMBER}
& \multicolumn{2}{c}{Object HalBench}
& \multicolumn{5}{c}{MME-Hallucination} \\
\cmidrule(lr){2-5} \cmidrule(lr){6-7} \cmidrule(lr){8-12}
& CH$\downarrow$ & Cov.$\uparrow$ & Hal$\downarrow$ & Cog$\downarrow$
& CH$_s\downarrow$ & CH$_i\downarrow$
& Exist.$\uparrow$ & Count$\uparrow$ & Pos.$\uparrow$ & Color$\uparrow$ & Total$\uparrow$ \\
\midrule

LLaVA-1.5-7B
& 8.3 & 45.0 & 32.0 & 2.2
& 27.0 & 10.5
& 180.00 & 110.00 & 121.33 & 141.67 & 553.00 \\

+ICD
& 6.3 & 46.3 & 25.8 & 2.2
& 22.3 & 7.4
& 180.00 & 126.67 & 113.33 & 143.33 & 563.33 \\

+OPERA
& \underline{4.6} & 45.6 & \underline{18.9} & 1.6
& 28.3 & 12.1
& \textbf{190.00} & 133.33 & 121.67 & 155.00 & 600.00 \\

+VCD
& 6.1 & 46.7 & 25.8 & 2.1
& 23.0 & 8.3
& \underline{185.00} & 126.67 & 121.67 & 135.00 & 568.33 \\

+M3ID
& 5.4 & \underline{48.7} & 25.5 & \underline{1.4}
& 26.3 & 9.2
& 173.33 & 106.67 & 96.67 & 155.00 & 531.67 \\

+AVISC
& 7.1 & 45.2 & 27.6 & 2.1
& 24.1 & 7.8
& 180.00 & 138.33 & 123.33 & 158.33 & 599.99 \\

+EVAS
& \textbf{4.7} & 46.1 & \textbf{18.8} & 1.6
& 27.0 & 11.3
& \textbf{190.00} & \underline{140.00} & 121.67 & 160.33 & 612.00 \\

+FuzzyCD
& 6.2 & 48.1 & 27.3 & 1.8
& \underline{20.3} & \underline{7.3}
& \underline{185.00} & \textbf{148.33} & 123.33 & 153.33 & 609.99 \\

+TAME
& 12.0 & 40.6 & 35.6 & 3.4
& 26.3 & 10.6
& \textbf{190.00} & 133.66 & \textbf{128.33} & \underline{163.00} & \textbf{614.99} \\

+YARD
& \underline{4.6} & \textbf{48.9} & 21.6 & \textbf{1.2}
& \textbf{19.7} & \textbf{6.5}
& \underline{185.00} & \underline{140.00} & \underline{125.00} & \textbf{163.33} & \underline{613.33} \\

\bottomrule
\end{tabular}
}
\end{table*}

\begin{table*}[t]
\centering
\caption{Discriminative hallucination evaluation on AMBER and POPE. Accuracy measures the proportion of correctly answered yes/no questions, while F1 balances precision and recall.
}
\vspace{-3pt}
\label{tab:discriminative_hallucination}
\small
\setlength{\tabcolsep}{5pt}
\renewcommand{\arraystretch}{1.1}
\begin{tabular}{lcc cccccccc}
\toprule
& \multicolumn{2}{c}{AMBER} & \multicolumn{8}{c}{POPE.MSCOCO} \\
\cmidrule(lr){2-3} \cmidrule(lr){4-11}
& \multicolumn{2}{c}{Discrimination}
& \multicolumn{2}{c}{Random}
& \multicolumn{2}{c}{Popular}
& \multicolumn{2}{c}{Adversarial}
& \multicolumn{2}{c}{ALL} \\
\cmidrule(lr){2-3} \cmidrule(lr){4-5} \cmidrule(lr){6-7} \cmidrule(lr){8-9} \cmidrule(lr){10-11}
Method & Accuracy & F1 & Accuracy & F1 & Accuracy & F1 & Accuracy & F1 & Accuracy & F1 \\
\midrule

LLaVA-1.5-7B \textcolor{blue} 
& 67.00 & 71.10 & 85.67 & 83.71 & 84.61 & 82.47 & 81.51 & 80.54 &83.93 & 82.24 \\

+ICD \textcolor{blue}
& 75.92 & 81.58 & 85.20 & 84.02 & 83.70 & 82.64 & 81.20 & 80.45 & 83.37& 82.37 \\

+VCD \textcolor{blue}
& 67.30 & 71.10 & 86.77 & 85.50 & 84.93 & 83.81 & 82.33 & 81.53 & 84.68 & 83.61 \\

+M3ID \textcolor{blue}
& 67.25 & 70.90 & 84.5 & 82.55 & 82.93 & 81.12 & 79.93 & 78.50 & 82.46 & 80.72 \\

+AVISC \textcolor{blue}
& 70.70 & 75.45 & 86.77 & 85.29 & 85.23 & 83.86 & 82.57 & 81.48 & 84.86 & 83.55 \\

+EVAS \textcolor{blue}
& 77.93 & 83.81 & 85.74 & 84.49 & 84.97 & 83.36 & 82.03 & 80.74 & 84.25 & 82.86 \\

+YARD
& \textbf{78.70} & \textbf{84.10} & \textbf{89.03} & \textbf{88.41} & \textbf{86.67} & \textbf{86.25} & \textbf{82.73} & \textbf{82.90} & \textbf{86.14} & \textbf{85.85} \\

\midrule

InstructBLIP \textcolor{blue}
& 68.20 & 74.60 & 82.61 & 82.47 & 79.51 & 79.59 & 78.98 & 79.43 & 80.37& 80.50 \\

+VCD \textcolor{blue}
& 69.65 & 75.90 & 86.01 &84.96 &83.33 & 82.49 & 81.03 & 80.57 & 83.49 & 82.68 \\

+AVISC \textcolor{blue}
& 72.60 & 78.60 & 85.67 & 84.31 & 83.57 & 82.41 & 80.90 & 80.14 & 83.38 & 82.29 \\

+YARD
& \textbf{73.62} & \textbf{79.07} & \textbf{90.77} & \textbf{90.05} & \textbf{89.07} & \textbf{88.43} & \textbf{87.33} & \textbf{86.87} & \textbf{89.06} & \textbf{88.45} \\

\bottomrule
\end{tabular}
\vspace{-5pt}
\end{table*}

\subsection{Main Results}

\paragraph{Does YARD reduce hallucinations more effectively?}
Tables~\ref{tab:hallucination_all} and~\ref{tab:discriminative_hallucination} compare YARD with existing hallucination mitigation methods.
Overall, YARD achieves consistent improvements on both generative and discriminative hallucination evaluations, indicating that the proposed degraded branch provides an effective contrastive signal.

On the generative benchmarks in Table~\ref{tab:hallucination_all}, YARD substantially reduces object-level hallucination.
On Object HalBench, it achieves the best CH$_s$ and CH$_i$, reducing them from 27.0/10.5 to 19.7/6.5 compared with the LLaVA-1.5-7B baseline.
YARD also improves AMBER hallucination metrics, lowering CH and Hal. while achieving the best coverage and cognition scores.
These results suggest that YARD suppresses hallucinated content without simply making the model overly conservative.
On MME-Hallucination, YARD remains competitive with the strongest baselines and obtains the best color score, showing that it preserves general visual perception ability.

Table~\ref{tab:discriminative_hallucination} shows that YARD also improves object-existence discrimination.
On LLaVA-1.5-7B, YARD improves AMBER accuracy/F1 from 67.00/71.10 to 78.70/84.10, and POPE accuracy/F1 from 83.93/82.24 to 86.14/85.85.
The same trend holds on InstructBLIP, where YARD improves POPE overall accuracy/F1 from 80.37/80.50 to 89.06/88.45.
This demonstrates that YARD is effective not only for reducing hallucinated descriptions, but also for discriminative hallucination evaluation.

Taken together, these results support our design hypothesis: an image-aware but locally under-grounded degraded branch provides a more targeted contrastive signal, suppressing predictions that are plausible under weak local grounding but unsupported by clean visual evidence.

\subsection{Transfer Across LVLM Architectures}

\paragraph{Does YARD generalize to different LVLM families?}
We further evaluate YARD as a plug-and-play decoding method across diverse LVLM architectures, including Qwen3-VL, Qwen2-VL, LLaVA-NeXT, LLaVA-v1.5, InstructBLIP, and Mini-Gemini. 
As shown in Table~\ref{tab:yard_chair_pope}, YARD consistently improves both Object HalBench and POPE across all evaluated models.

The gains are observed on both strong and hallucination-prone backbones. 
For example, on Qwen3-VL-4B, YARD improves POPE F1/Acc. from 89.2/89.8 to 92.7/93.6, while reducing CH$_s$/CH$_i$ from 8.7/6.1 to 7.5/5.4. 
On LLaVA-NeXT, where object hallucination is more severe, YARD reduces CH$_s$/CH$_i$ from 20.5/9.3 to 11.0/5.2.

These consistent improvements suggest that YARD is not tied to a specific LLaVA-style architecture. 
Instead, middle-layer register-conditioned degradation captures a general property of LVLM decoding, enabling training-free transfer across different model families.

\begin{figure}[h]
\centering
\captionsetup{font=normalsize}

\begin{tcolorbox}[
  enhanced,
  colback=white,
  colframe=yardblue,
  boxrule=0.7pt,
  arc=2pt,
  left=4pt,
  right=4pt,
  top=4pt,
  bottom=4pt,
  width=\columnwidth,
  before skip=0pt,
  after skip=0pt
]

\small
\setlength{\parskip}{0pt}
\renewcommand{\baselinestretch}{0.92}\selectfont

\includegraphics[
  width=\linewidth,
  height=0.2\textheight,
  keepaspectratio
]{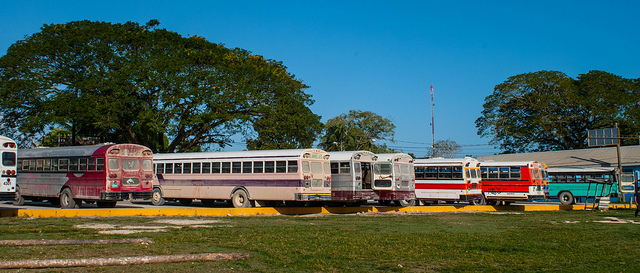}

\begin{casebox}
\textbf{Q:} Please describe this image in detail.
\end{casebox}

\begin{casebox}
\textcolor{yardorange}{\textbf{Baseline:}}
The image conveys the story of the two school buses, one red and blue, parked next to each other in the grass or dirt.
\hall{There's a \textbf{boy} near the blue bus on the right, possibly disembarking or about to board either bus.}
\hall{Additionally, a single \textbf{car} and a row of parked buses are visible in the open area.}
\end{casebox}

\begin{casebox}
\textcolor{yardorange}{\textbf{YARD:}}
The image shows buses parked next to each other, lined up on a grassy area near a parking lot.
The buses appear to be in various states of maintenance and come in various colors.
Large trees and a clear blue sky can be seen in the background, giving the scene an open outdoor setting.
\end{casebox}

\end{tcolorbox}
\vspace{3pt}
\caption{
Qualitative comparison between baseline and \textbf{Yard}.
Hallucinated sentences are highlighted in red.
}
\vspace{-15pt}
\label{fig:bus_case}
\vspace{-8pt}
\end{figure}

\subsection{Qualitative Analysis}

\paragraph{Are YARD responses more visually grounded?}
Figure~\ref{fig:bus_case} shows a representative open-ended generation example.
The baseline response introduces unsupported objects, such as a boy near the bus and a single car, which are plausible in the scene but not grounded in the image.
In contrast, YARD focuses on visually supported content, including the buses, grassy area, trees, and sky, without adding non-existent objects.

This example illustrates how YARD suppresses language-plausible but visually unsupported predictions.
By contrasting the clean branch with an image-aware but locally under-grounded degraded branch, YARD encourages responses that remain more faithful to the visual evidence.

\subsection{Ablation Study}

\paragraph{Are middle-layer branching and register-conditioned degradation necessary?}
Table~\ref{tab:yard_ablation} ablates two key design choices of YARD: the branching layer $K$ and the degraded visual condition.

First, the branching layer is critical.
Branching too early ($K=1$) provides insufficient multimodal context, leading to weaker POPE performance and higher CHAIR scores.
Branching too late ($K=25$) is also suboptimal, as visual evidence has been largely absorbed into text-side representations, leaving limited room for degradation.
In contrast, middle-layer branching ($K=10$) achieves the best overall performance, supporting our observation that this stage is the critical window for local visual evidence extraction.

Second, the degraded condition also matters.
Average degradation remains inferior to YARD, suggesting that coarse averaged features are not sufficiently informative.
Text-only variants remove visual information more aggressively and provide a less image-specific contrastive signal; although they reduce CHAIR to some extent, they yield weaker POPE Acc./F1.
Overall, YARD achieves the best POPE Acc./F1 and the lowest CH$_s$/CH$_i$, confirming that middle-layer branching and register-conditioned degradation are both necessary for constructing an image-aware but locally under-grounded degraded branch.

\begin{table}[t]
\centering
\caption{Cross-architecture evaluation of YARD on Object HalBench and POPE across diverse VLM backbones.}
\label{tab:yard_chair_pope}
\small
\setlength{\tabcolsep}{5pt}
\renewcommand{\arraystretch}{1.08}
\begin{tabular}{lcccc}
\toprule
\textbf{Model} 
& \multicolumn{2}{c}{\textbf{Object HalBench}} 
& \multicolumn{2}{c}{\textbf{POPE}} \\
\cmidrule(lr){2-3} \cmidrule(lr){4-5}
& \textbf{$\mathrm{CH}_{s}\downarrow$} 
& \textbf{$\mathrm{CH}_{i}\downarrow$} 
& \textbf{F1$\uparrow$} 
& \textbf{Acc.$\uparrow$} \\
\midrule

Qwen3-VL-4B & 8.7 & 6.1 & 89.2 & 89.8 \\
\rowcolor{gray!10}
\quad + YARD 
& \textbf{7.5}
& \textbf{5.4}
& \textbf{92.7}
& \textbf{93.6} \\

Qwen3-VL-8B & 7.9 & 5.4 & 88.4 & 88.9 \\
\rowcolor{gray!10}
\quad + YARD 
& \textbf{7.3}
& \textbf{5.0}
& \textbf{91.2}
& \textbf{91.7} \\

Qwen2-VL-7B & 12.7 & 7.8 & 87.1 & 87.9 \\
\rowcolor{gray!10}
\quad + YARD 
& \textbf{10.8}
& \textbf{6.3}
& \textbf{88.6}
& \textbf{89.2} \\

LLaVA-NeXT & 20.5 & 9.3 & 83.1 & 84.7 \\
\rowcolor{gray!10}
\quad + YARD 
& \textbf{11.0}
& \textbf{5.2}
& \textbf{86.5}
& \textbf{87.3} \\

LLaVA-v1.5 & 27.0 & 10.5 & 82.2 & 83.9 \\
\rowcolor{gray!10}
\quad + YARD 
& \textbf{19.7}
& \textbf{6.5}
& \textbf{85.9}
& \textbf{86.1} \\

InstructBLIP & 15.3 & 5.5 & 80.5 & 80.4 \\
\rowcolor{gray!10}
\quad + YARD 
& \textbf{12.5}
& \textbf{4.2}
& \textbf{88.5}
& \textbf{89.1} \\

Mini-Gemini & 14.0 & 6.9 & 85.1 & 86.6 \\
\rowcolor{gray!10}
\quad + YARD 
& \textbf{13.2}
& \textbf{6.5}
& \textbf{86.4}
& \textbf{87.1} \\

\bottomrule
\end{tabular}
\vspace{-10pt}
\end{table}

\begin{table}[h]
\centering
\small
\setlength{\tabcolsep}{5.8pt}
\renewcommand{\arraystretch}{1.12}
\caption{
Ablation study of degraded-branch construction. 
$K$ denotes the branching layer; Average uses averaged visual features, and Text-Only removes visual tokens. 
Acc. and F1 are averaged over POPE subsets; lower CH$_s$/CH$_i$ is better.
}
\label{tab:yard_ablation}
\begin{tabular}{lcccc}
\toprule
\multirow{2}{*}{Variant} 
& \multicolumn{2}{c}{POPE} 
& \multicolumn{2}{c}{CHAIR} \\
\cmidrule(lr){2-3} \cmidrule(lr){4-5}
& Acc.$\uparrow$ & F1$\uparrow$ & CH$_s\downarrow$ & CH$_i\downarrow$ \\
\midrule
YARD ($K=1$)              & 84.5 & 83.7 & 22.9 & 7.3 \\
YARD ($K=25$)             & 83.2 & 81.6 & 25.0 & 7.9 \\
Average. ($K=10$)& 84.9 & \underline{84.3} & 23.7 & 7.7 \\
Text-Only. ($K=1$)    & 84.3 & 82.6 & 23.0 & 7.2 \\
Text-Only. ($K=10$)   & \underline{85.1} & 84.2 & \underline{21.7} & \underline{7.0} \\
\midrule
YARD ($K=10$)             & \textbf{86.1} & \textbf{85.9} & \textbf{19.7} & \textbf{6.5} \\
\bottomrule
\end{tabular}
\vspace{-10pt}
\end{table}

\section{Conclusion}

We propose YARD, a training-free Y-Architecture contrastive decoding framework. YARD introduces register-based feature-level degradation at the middle decoder layers to construct an image-aware but locally under-grounded degraded branch, while sharing shallow-layer computation to avoid a complete second forward pass. Experiments show that YARD consistently reduces hallucinations across multiple LVLM architectures and benchmarks.

\section*{Limitations}

YARD is training-free at the decoding stage, but it relies on constructing effective register representations for the target LVLM. While this design is effective for the Transformer-based LVLMs evaluated in this work, architectures with different vision encoders, LLM backbones, or vision-language projection mechanisms may require additional adaptation. In addition, the performance of YARD may be affected by the choice of branching layer and the construction of register representations. Extending register-based degradation to broader multimodal architectures and input formats is a natural direction for future work.

\nocite{*}
\bibliography{custom}
\clearpage
\appendix

\section{Appendix}
\label{sec:appendix}

\paragraph{Overview.}
This appendix provides additional analyses and implementation details for \textbf{YARD}. 
We first report the inference-time comparison~\ref{effiency} to analyze the hallucination--efficiency trade-off, followed by the implementation details~\ref{app:implementation_details} used across experiments. 
We then provide distributional analyses of clean and degraded branches, including both aggregate divergence metrics~\ref{app:distributional_analysis} and a top-token case study~\ref{app:top_token_case}, to further explain why register-conditioned feature-level degradation forms a more suitable degraded branch than pixel-level or text-only alternatives. 
Next, we examine the sensitivity of \textbf{YARD} to the number of register tokens and the contrastive strength $\alpha$~\ref{app:param_sensitivity}. 
We also evaluate YARD on general multimodal benchmarks~\ref{app:general_benchmarks} to verify that hallucination mitigation is not achieved by sacrificing general visual understanding. 
Finally, we describe the evaluation benchmarks~\ref{app:benchmark_details}, present a detailed formulation of YARD and contrastive decoding~\ref{app:yard_formulation}, and provide additional qualitative examples.

\subsection{Efficiency Analysis}
\label{effiency}
\begin{figure}[h]
\centering
\includegraphics[width=\columnwidth]{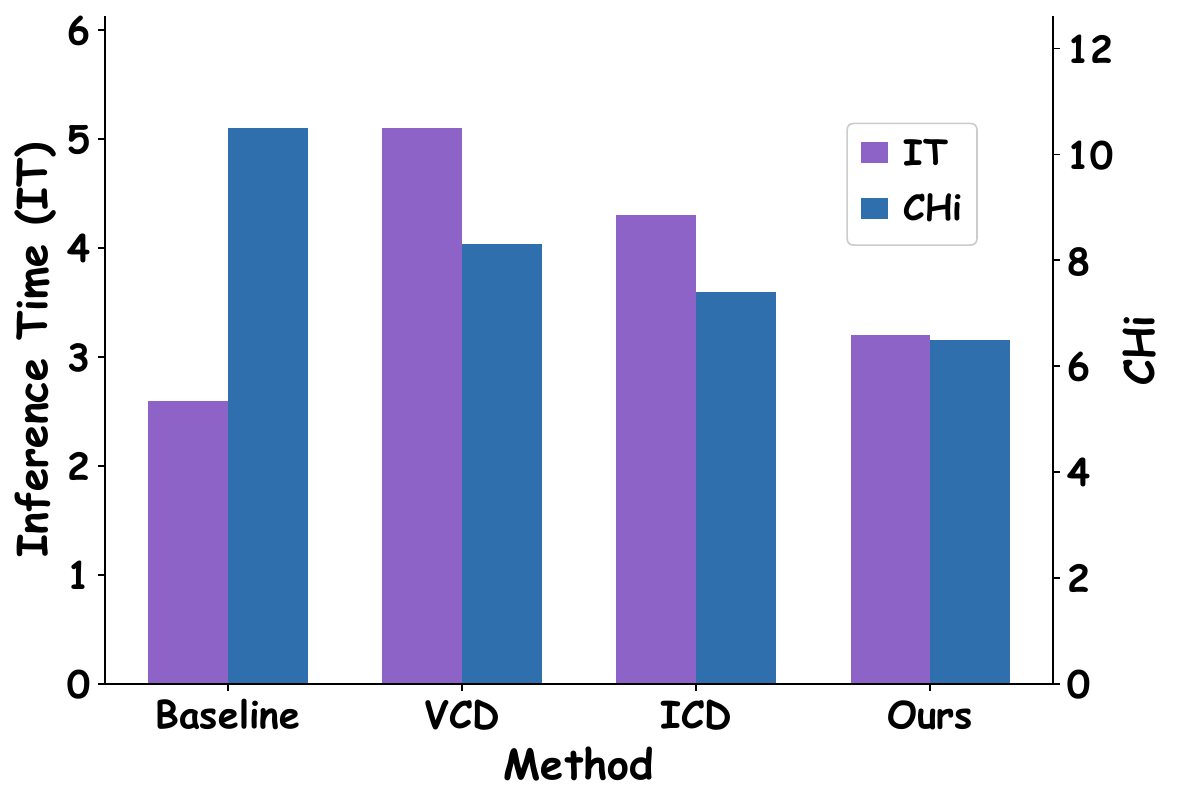}
\caption{
Comparison of inference time and hallucination rate across decoding methods.
YARD achieves lower CH$_i$ with only a modest inference overhead compared with the baseline, while prior contrastive decoding methods introduce substantially higher cost or hallucination.
}
\label{fig:time_chi}
\end{figure}

We further analyze the inference cost of YARD compared with representative decoding-time hallucination mitigation methods.
Figure~\ref{fig:time_chi} reports inference time (IT) and CH$_i$ on LLaVA-1.5-7B~\citep{liu2024improved}.
The vanilla baseline requires 2.6 seconds and obtains a CH$_i$ of 10.5.
VCD~\citep{leng2024mitigating} reduces CH$_i$ to 8.3, but requires an additional degraded forward pass, increasing inference time to 5.1 seconds.
ICD~\citep{wang2024mitigating} further reduces CH$_i$ to 7.4, but still incurs a substantial inference overhead, requiring 4.3 seconds.

In contrast, YARD achieves the lowest CH$_i$ of 6.5 with only a modest increase in inference time from 2.6 to 3.2 seconds.
This efficiency comes from the Y-Architecture design: the clean and degraded branches share the shallow decoder layers, and the degraded branch is only computed after the middle-layer split.
Thus, YARD provides a stronger hallucination--efficiency trade-off than prior contrastive decoding methods that rely on largely separate degraded computation.

\subsection{Implementation Details}
\label{app:implementation_details}

YARD is applied in a fully training-free manner without parameter updates, additional supervision, or external verifiers. 
Unless otherwise specified, we construct the degraded branch at approximately one-third of the LLM decoder depth, following our observation that text-to-vision grounding begins to emerge in the middle layers. 
For LLaVA-1.5-7B\cite{liu2024improved}, this corresponds to branching after the 10-th decoder layer. 
For other backbones, we use the same one-third-depth rule to select the branching layer.

We use 1 register token to construct the degraded visual condition in all experiments. 
The contrastive strength is set to $\alpha=1$ by default. 
These hyperparameters are kept fixed across benchmarks unless explicitly stated otherwise. 
All evaluations are conducted with the original model weights and the default decoding settings of each backbone. All reported results are consistently reproduced across multiple independent runs, indicating that the observed improvements are stable rather than incidental.

\definecolor{lightred}{RGB}{255,238,238}

\begin{table}[h]
\centering
\scriptsize
\setlength{\tabcolsep}{3.2pt}
\renewcommand{\arraystretch}{1.08}
\caption{
Distributional gap between clean and degraded branches.
Top-1 Match measures whether the two branches predict the same top token.
KL, symmetric KL, JS, and TV measure distributional divergence between clean and degraded output distributions.
}
\label{tab:clean_degraded_gap}
\begin{tabular}{lccccc}
\toprule
Method 
& Top-1 Match$\downarrow$
& KL$\uparrow$
& Sym. KL$\uparrow$
& JS$\uparrow$
& TV$\uparrow$ \\
\midrule
Pixel-level 
& 0.9233 & 0.0617 & 0.1246 & 0.0130 & 0.0682 \\

\rowcolor{lightred}
Ours-Reg. 
& 0.7121 & 0.7380 & 1.6726 & 0.1197 & 0.2740 \\

Text-only 
& 0.6675 & 0.8667 & 2.0323 & 0.1406 & 0.3067 \\
\bottomrule
\end{tabular}
\end{table}

\subsection{Distributional Analysis of Clean and Degraded Branches}
\label{app:distributional_analysis}

To further understand the behavior of different degraded-branch constructions, we compare the output distributions of the clean and degraded branches in Table~\ref{tab:clean_degraded_gap}.
Top-1 Match measures how often the two branches predict the same most likely token, while KL, symmetric KL, JS, and TV quantify the distributional discrepancy between the two output distributions.
We use these metrics as diagnostic tools rather than direct performance measures: an effective degraded branch should be different enough from the clean branch to expose hallucination-prone predictions, but should not be so distant that it becomes unrelated to the current image.

Pixel-level degradation remains too close to the clean branch.
It obtains a high Top-1 Match of 0.9233 and very small divergence values across KL, JS, and TV.
This suggests that the pixel-level degraded branch still preserves substantial normal visual semantics and correct object evidence from the clean branch.
Consequently, when the clean-logit anchoring term is removed in the w/o $\alpha$ stress test, direct subtraction may suppress not only hallucination-related logits but also visually grounded predictions, leading to a large increase in CH$_s$ and CH$_i$ in Table~\ref{tab:degradation_alpha}.

Text-only degradation exhibits the opposite behavior.
It has the lowest Top-1 Match and the largest distributional divergences, indicating that its degraded logits are much farther from the clean branch.
Although this strong shift can induce hallucination-prone predictions, the resulting distribution is largely image-agnostic and dominated by language priors.
Therefore, it is less aligned with image-specific hallucination components in the clean branch and does not faithfully model the case where the model has seen the image but fails to use local visual evidence correctly.

Feature-level degradation with register tokens yields a more balanced distributional gap.
Compared with pixel-level degradation, it produces a much larger discrepancy from the clean branch, suggesting that local grounding has been effectively weakened.
Compared with text-only degradation, it remains closer to the clean branch, indicating that it still preserves image-level semantics rather than collapsing into a pure language prior.
This intermediate behavior matches our intended degraded condition: image-aware but locally under-grounded.
Together with the w/o $\alpha$ stress test, these results support that register-conditioned feature-level degradation provides a more targeted contrastive signal for suppressing visually unsupported predictions.

\definecolor{lightpurple}{RGB}{238,235,255}

\begin{table}[h]
\centering
\small
\setlength{\tabcolsep}{5.2pt}
\renewcommand{\arraystretch}{1.12}
\caption{
Sensitivity analysis on the number of register tokens. 
The shaded column denotes the default setting used in our experiments. 
Higher F1 is better, while lower CH$_s$ and CH$_i$ indicate fewer hallucinations.
}
\label{tab:reg_num_sensitivity}
\begin{tabular}{lcccccc}
\toprule
\# Reg. Tokens & \cellcolor{lightpurple}1 & 2 & 4 & 8 & 16 & 32 \\
\midrule
F1$\uparrow$ 
& \cellcolor{lightpurple}\textbf{85.9} & \underline{86.0} & 85.9 & \textbf{86.1} & \underline{86.0} & 85.7 \\
CH$_s\downarrow$ 
& \cellcolor{lightpurple}\underline{19.7} & 20.3 & 21.7 & 20.7 & \textbf{19.3} & 21.3 \\
CH$_i\downarrow$ 
& \cellcolor{lightpurple}\underline{6.5} & \underline{6.5} & 7.0 & 7.1 & \textbf{6.2} & 6.7 \\
\bottomrule
\end{tabular}
\end{table}

\subsection{Parameter Sensitivity Analysis}
\label{app:param_sensitivity}

\paragraph{Number of register tokens.}
Table~\ref{tab:reg_num_sensitivity} studies the effect of using different numbers of register tokens in the degraded branch.
Overall, YARD is relatively robust to this hyperparameter: across 1 to 32 register tokens, POPE F1\cite{li2023evaluating} remains within a narrow range from 85.7 to 86.1, while CH$_s$ and CH$_i$ also remain consistently low.
Using too few register tokens may provide insufficient capacity to represent global image-level semantics, whereas using too many registers can introduce redundant degraded visual conditions and slightly weaken the contrastive signal.
We choose 1 register token as the default setting.

\definecolor{lightpurple}{RGB}{238,235,255}

\begin{table}[h]
\centering
\small
\setlength{\tabcolsep}{6.2pt}
\renewcommand{\arraystretch}{1.12}
\caption{
Sensitivity analysis on the contrastive strength $\alpha$. 
The shaded column denotes the default setting used in our experiments. 
Higher F1 is better, while lower CH$_s$ and CH$_i$ indicate fewer hallucinations.
}
\label{tab:alpha_sensitivity}
\begin{tabular}{lccccc}
\toprule
$\alpha$ & w/o & 0.25 & 0.5 & 0.75 & \cellcolor{lightpurple}1.0 \\
\midrule
F1$\uparrow$ 
& 85.0 & 85.2 & 85.5 & \underline{85.9} & \cellcolor{lightpurple}\textbf{85.9} \\
CH$_s\downarrow$ 
& 21.0 & 20.3 & \textbf{17.7} & \underline{18.0} & \cellcolor{lightpurple}19.7 \\
CH$_i\downarrow$ 
& 7.0 & 7.0 & 6.6 & \textbf{5.9} & \cellcolor{lightpurple}\underline{6.5} \\
\bottomrule
\end{tabular}
\end{table}

\paragraph{Contrastive strength $\alpha$.}
Table~\ref{tab:alpha_sensitivity} analyzes the sensitivity of YARD to the contrastive strength $\alpha$.
As $\alpha$ increases, the degraded-branch signal is more strongly subtracted, which generally improves both POPE F1\cite{li2023evaluating} and hallucination metrics.
For example, increasing $\alpha$ from 0 to 1.0 improves POPE F1\cite{li2023evaluating} from 85.0 to 85.9 and reduces CH$_i$ from 7.0 to 6.5.

We use $\alpha=1.0$ as the default setting, as it offers a stable trade-off between discriminative accuracy and hallucination reduction.

\begin{table}[h]
\centering
\small
\setlength{\tabcolsep}{6.5pt}
\renewcommand{\arraystretch}{1.10}
\caption{
General multimodal benchmark results on LLaVA-1.5-7B. 
YARD preserves general visual understanding performance while improving hallucination robustness. 
Higher values indicate better performance.
}
\label{tab:general_benchmarks}
\begin{tabular}{lccc}
\toprule
Benchmark & Baseline & YARD & $\Delta$ \\
\midrule
GQA      & 61.9 & \textbf{62.1} & +0.2 \\
MMBench  & 64.7 & \textbf{64.9} & +0.2 \\
TextVQA  & \textbf{58.2} & \textbf{58.2} & 0.0 \\
SQA$^{\mathrm{I}}$ & 69.5 & \textbf{69.6} & +0.1 \\
MME      & 1594 & \textbf{1741} & +147 \\
\bottomrule
\end{tabular}
\end{table}

\subsection{General Multimodal Benchmark Results}
\label{app:general_benchmarks}

Hallucination mitigation methods may reduce unsupported generations by making the model overly conservative or by weakening its general visual understanding ability.
To examine whether YARD introduces such a trade-off, we additionally evaluate it on standard multimodal benchmarks that are not specifically designed for hallucination measurement.
These benchmarks cover different aspects of general LVLM capability, including visual question answering, text-rich visual understanding, science question answering, and comprehensive multimodal perception.

As shown in Table~\ref{tab:general_benchmarks}, YARD largely preserves the general multimodal performance of the base LLaVA-1.5-7B\cite{liu2024improved} model.
On GQA\cite{hudson2019gqa}, MMBench\cite{liu2024mmbench}, and SQA$^{\mathrm{I}}$, YARD achieves small but consistent improvements over the baseline.
On TextVQA\cite{singh2019towards}, where fine-grained text recognition and visual-linguistic alignment are important, YARD maintains the same performance as the baseline.
These results suggest that the proposed register-conditioned degraded branch does not simply suppress visual information or make the model less responsive to image content.

Notably, YARD improves the MME\cite{fu2026mme} score from 1594 to 1741.
Since MME evaluates a broad range of perception and cognition abilities, this improvement indicates that YARD can enhance hallucination robustness while preserving, and in some cases improving, general visual perception.
Together with the hallucination-specific results in the main paper, these findings suggest that YARD provides a targeted contrastive signal: it suppresses visually unsupported predictions without broadly degrading the model's multimodal understanding ability.

\subsection{Benchmark Details}
\label{app:benchmark_details}

\paragraph{AMBER\cite{wang2023amber}.}
AMBER is a hallucination-oriented benchmark for evaluating both generative and discriminative hallucination in LVLMs. 
For generative evaluation, it reports hallucination-related metrics such as CH, Hal., and Cog., together with coverage, which reflects whether the model preserves sufficient visual content. 
For discriminative evaluation, it measures whether the model correctly judges visual content. 
We report both generative metrics and discrimination accuracy/F1.

\paragraph{Object HalBench / CHAIR\cite{yu2024rlhf}.}
Object HalBench evaluates object hallucination in image captioning and open-ended generation. 
It measures whether generated object mentions are supported by the image. 
We report CH$_s$ and CH$_i$, where lower values indicate fewer hallucinated objects at the sentence and instance levels. 
This benchmark directly reflects whether YARD suppresses visually unsupported object predictions.

\paragraph{MME-Hallucination\cite{fu2026mme}.}
MME-Hallucination evaluates fine-grained hallucination-related perception abilities, including object existence, count, position, and color. 
These categories test whether the model can produce visually faithful responses under different grounding requirements. 
We report each subcategory score and the total score, where higher values indicate better performance.

\paragraph{POPE-MSCOCO\cite{li2023evaluating}.}
POPE-MSCOCO evaluates object hallucination through binary object-existence questions. 
It contains random, popular, and adversarial sampling settings, where the adversarial split includes objects more likely to trigger language-prior hallucination. 
We report accuracy and F1 for each split and the overall result.

\paragraph{GQA\cite{hudson2019gqa}.}
GQA is a visual question answering benchmark emphasizing real-world visual reasoning and compositional understanding. 
It covers objects, attributes, spatial relations, and multi-step reasoning. 
We use GQA to evaluate whether YARD preserves general visual reasoning beyond hallucination-specific settings.

\paragraph{MMBench\cite{liu2024mmbench}.}
MMBench evaluates general multimodal understanding across perception and reasoning categories. 
It contains multiple-choice questions covering diverse visual and linguistic abilities. 
We report overall accuracy to assess broad multimodal capability.

\paragraph{TextVQA\cite{singh2019towards}.}
TextVQA evaluates visual question answering over text-rich images. 
It requires recognizing and reasoning over scene text, making it sensitive to fine-grained visual-text alignment. 
We use it to examine whether YARD preserves text-oriented visual understanding.

\paragraph{ScienceQA-IMG\cite{lu2022learn}.}
ScienceQA-IMG is the image subset of ScienceQA. 
It covers science questions requiring both visual understanding and textual reasoning. 
We report accuracy on this subset to evaluate multimodal reasoning in scientific contexts.

\paragraph{MME\cite{fu2026mme}.}
MME is a comprehensive benchmark for perceptual and cognitive abilities of multimodal large language models. 
It covers object recognition, counting, spatial reasoning, color perception, OCR, and commonsense reasoning. 
We report the total MME score to assess whether YARD preserves general multimodal perception while mitigating hallucinations.

\subsection{Detailed Formulation of YARD}
\label{app:yard_formulation}

This section provides a more detailed formulation of YARD. 
Given an input image $I$ and a text query $x$, the vision encoder first produces a sequence of patch-level visual tokens
\begin{equation}
\mathbf{Z} = [\mathbf{z}_1,\ldots,\mathbf{z}_N] \in \mathbb{R}^{N \times d_v},
\end{equation}
where each $\mathbf{z}_j$ corresponds to a local image patch. 
The visual tokens are then mapped into the LLM hidden space through the multimodal projector $\phi(\cdot)$:
\begin{equation}
\tilde{\mathbf{Z}} = \phi(\mathbf{Z}) \in \mathbb{R}^{N \times d}.
\end{equation}
For simplicity, we use $\mathbf{Z}$ to denote the projected visual tokens in the following derivation.

\paragraph{Training-free register construction.}
YARD constructs register tokens in a training-free manner. 
Following test-time register construction, we append $M$ non-image register tokens to the visual sequence:
\begin{equation}
\mathbf{R}^{(0)} = [\mathbf{r}^{(0)}_1,\ldots,\mathbf{r}^{(0)}_M] \in \mathbb{R}^{M \times d}.
\end{equation}
These tokens are not associated with any spatial image patch. 
Instead, they are used to absorb global image-level information that would otherwise be stored in high-norm outlier patch tokens.

Let $\mathcal{K}_{\mathrm{reg}}$ denote the set of register neurons identified in the vision encoder. 
For a selected MLP layer, let $a_{j,k}$ denote the activation of neuron $k$ on patch token $j$. 
For each register neuron $k \in \mathcal{K}_{\mathrm{reg}}$, we redirect the maximum anomalous activation from patch tokens to the appended register token:
\begin{equation}
\begin{aligned}
a_{r,k} &\leftarrow \max_{1 \leq j \leq N} a_{j,k}, \\
a_{j,k} &\leftarrow 0, \quad \forall j \in \{1,\ldots,N\}.
\end{aligned}
\end{equation}
This operation shifts high-norm activation away from image patch tokens and into the register tokens, yielding register representations that preserve coarse image-level semantics while lacking reliable patch-level correspondence. 
The final LLM input sequence is therefore
\begin{equation}
\mathbf{X}^{(0)} = [\mathbf{Z};\mathbf{R};\mathbf{T}],
\end{equation}
where $\mathbf{T}=[\mathbf{t}_1,\ldots,\mathbf{t}_L]$ denotes the text-token sequence.

\paragraph{Shared-prefix decoding.}
Let the LLM decoder contain $D$ layers, denoted as
\begin{equation}
F_{1:D} = F_D \circ F_{D-1} \circ \cdots \circ F_1 .
\end{equation}
YARD chooses a branching layer $K$ around the middle decoder layers, where text-to-vision grounding begins to emerge. 
Before branching, the clean and degraded branches share the same computation:
\begin{equation}
\mathbf{H}^{(K)}
=
F_{1:K}(\mathbf{X}^{(0)})
=
[\mathbf{Z}^{(K)};\mathbf{R}^{(K)};\mathbf{T}^{(K)}].
\end{equation}
This shared prefix serves two purposes. 
First, it allows both branches to inherit the same early multimodal context. 
Second, it avoids recomputing the shallow decoder layers, reducing the overhead compared with input-level contrastive decoding.

\paragraph{Y-Architecture branch construction.}
At layer $K$, YARD splits the computation into a clean branch and a degraded branch:
\begin{equation}
\mathbf{H}^{(K)}_c
=
[\mathbf{Z}^{(K)};\mathbf{R}^{(K)};\mathbf{T}^{(K)}],
\end{equation}
\begin{equation}
\mathbf{H}^{(K)}_d
=
[\mathbf{R}^{(K)};\mathbf{T}^{(K)}].
\end{equation}
The clean branch retains the original patch-level visual tokens $\mathbf{Z}^{(K)}$ and therefore preserves access to fine-grained local visual evidence. 
In contrast, the degraded branch removes $\mathbf{Z}^{(K)}$ and relies only on register tokens $\mathbf{R}^{(K)}$ as the visual condition. 
Since register tokens contain global image semantics but lack stable local correspondence, the degraded branch remains image-aware while being locally under-grounded.

The two branches are then forwarded through the remaining decoder layers:
\begin{equation}
\begin{aligned}
\mathbf{H}^{(D)}_c &= F_{K+1:D}\!\left(\mathbf{H}^{(K)}_c\right), \\
\mathbf{H}^{(D)}_d &= F_{K+1:D}\!\left(\mathbf{H}^{(K)}_d\right).
\end{aligned}
\end{equation}
Let $\mathbf{h}^{(D)}_{c,i}$ and $\mathbf{h}^{(D)}_{d,i}$ denote the hidden states used for next-token prediction at decoding step $i$ in the clean and degraded branches. 
The corresponding vocabulary-level logits are
\begin{equation}
\boldsymbol{\ell}^{c}_{i}
=
g_{\mathrm{LM}}(\mathbf{h}^{(D)}_{c,i}),
\qquad
\boldsymbol{\ell}^{d}_{i}
=
g_{\mathrm{LM}}(\mathbf{h}^{(D)}_{d,i}),
\end{equation}
where $g_{\mathrm{LM}}(\cdot)$ is the language modeling head.

\paragraph{Register-conditioned contrastive decoding.}
YARD combines the clean and degraded logits using the standard contrastive decoding form:
\begin{equation}
\boldsymbol{\ell}^{\mathrm{yard}}_{i}
=
(1+\alpha)\boldsymbol{\ell}^{c}_{i}
-
\alpha \boldsymbol{\ell}^{d}_{i},
\end{equation}
where $\alpha$ controls the strength of contrastive subtraction. 
The next-token distribution is then computed as
\begin{equation}
p_{\mathrm{yard}}(y_i \mid I,x,y_{<i})
=
\mathrm{softmax}
\left(
\boldsymbol{\ell}^{\mathrm{yard}}_{i}
\right)_{y_i}.
\end{equation}

Following common contrastive decoding practice, we apply a clean-branch plausibility constraint to avoid promoting tokens that are unlikely under the clean branch. 
Let
\begin{equation}
p_c(y)
=
\mathrm{softmax}(\boldsymbol{\ell}^{c}_{i})_y
\end{equation}
be the clean-branch probability for token $y$. 
We define the candidate set
\begin{equation}
\mathcal{V}_i
=
\left\{
y \in \mathcal{V}
\mid
p_c(y) \geq \tau \max_{y' \in \mathcal{V}} p_c(y')
\right\},
\end{equation}
where $\tau$ is a plausibility threshold and $\mathcal{V}$ is the vocabulary. 
The final distribution is restricted to $\mathcal{V}_i$:
\begin{equation}
p_{\mathrm{yard}}(y_i)
=
\frac{
\exp(\ell^{\mathrm{yard}}_{i,y_i})
\mathbb{I}[y_i \in \mathcal{V}_i]
}{
\sum_{y \in \mathcal{V}_i}
\exp(\ell^{\mathrm{yard}}_{i,y})
}.
\end{equation}

\paragraph{Dual degradation effect.}
YARD creates a dual degradation mechanism. 
The first degradation comes from the branching position. 
Since the degraded branch splits at layer $K$, it only inherits early-to-middle multimodal representations:
\begin{equation}
\mathbf{H}^{(K)} = [\mathbf{Z}^{(K)};\mathbf{R}^{(K)};\mathbf{T}^{(K)}],
\end{equation}
where text-to-vision grounding has begun to emerge but fine-grained local evidence has not yet been fully integrated into text-side representations. 
The second degradation comes from the visual condition after branching:
\begin{equation}
[\mathbf{Z}^{(K)};\mathbf{R}^{(K)}]
\quad \longrightarrow \quad
[\mathbf{R}^{(K)}].
\end{equation}
This removes patch-level local evidence while retaining register-level global semantics. 
As a result, the degraded logits tend to emphasize predictions that are plausible under coarse image semantics but insufficiently supported by local visual grounding. 
Contrastive subtraction then suppresses such predictions:
\begin{equation}
\ell^{\mathrm{yard}}_{i,y}
=
\ell^{c}_{i,y}
+
\alpha
\left(
\ell^{c}_{i,y}
-
\ell^{d}_{i,y}
\right).
\end{equation}
If a token $y$ is overly favored by the locally under-grounded degraded branch, i.e.,
\begin{equation}
\ell^{d}_{i,y} \gg \ell^{c}_{i,y},
\end{equation}
its final logit is reduced. 
This is the mechanism by which YARD suppresses language-plausible but visually unsupported hallucination-prone tokens.

\paragraph{Computational advantage.}
Compared with input-level contrastive decoding, which requires two largely independent forward passes,
\begin{equation}
\mathrm{Cost}_{\mathrm{input\text{-}CD}}
\approx
2 \cdot \mathrm{Cost}(F_{1:D}),
\end{equation}
YARD shares the prefix computation and only duplicates the remaining layers:
\begin{equation}
\begin{aligned}
\mathrm{Cost}_{\mathrm{YARD}}
\approx\;&
\mathrm{Cost}(F_{1:K}) \\
&+ \mathrm{Cost}\!\left(F_{K+1:D}; |\mathbf{Z}|+|\mathbf{R}|+|\mathbf{T}|\right) \\
&+ \mathrm{Cost}\!\left(F_{K+1:D}; |\mathbf{R}|+|\mathbf{T}|\right).
\end{aligned}
\end{equation}
Since the degraded branch removes the $N$ patch-level visual tokens after layer $K$, its remaining sequence length is shorter:
\begin{equation}
|\mathbf{R}| + |\mathbf{T}|
\ll
|\mathbf{Z}| + |\mathbf{R}| + |\mathbf{T}|.
\end{equation}
Thus, YARD improves efficiency by both sharing shallow-layer computation and reducing the sequence length of the degraded branch.

\paragraph{Contrastive decoding.}
Contrastive decoding is an inference-time strategy for mitigating hallucinations by contrasting the model prediction under a clean visual condition with that under a degraded condition.
Given an image $v$, a text query $t$, and previously generated tokens $y_{<i}$ at decoding step $i$, an LVLM produces vocabulary-level logits conditioned on the original multimodal input:
\begin{equation}
\boldsymbol{\ell}^{c}_{i}
=
f_{\theta}(v,t,y_{<i}) \in \mathbb{R}^{|\mathcal{V}|},
\end{equation}
where $\mathcal{V}$ denotes the vocabulary and the superscript $c$ indicates the clean branch.
The corresponding clean next-token distribution is
\begin{equation}
p^{c}_{i}(y)
=
\frac{\exp(\boldsymbol{\ell}^{c}_{i,y})}
{\sum_{y' \in \mathcal{V}} \exp(\boldsymbol{\ell}^{c}_{i,y'})}.
\end{equation}

To obtain a hallucination-prone reference, contrastive decoding further constructs a degraded condition $v_d$ and computes the degraded logits:
\begin{equation}
\boldsymbol{\ell}^{d}_{i}
=
f_{\theta}(v_d,t,y_{<i}) \in \mathbb{R}^{|\mathcal{V}|},
\end{equation}
with the corresponding degraded distribution
\begin{equation}
p^{d}_{i}(y)
=
\frac{\exp(\boldsymbol{\ell}^{d}_{i,y})}
{\sum_{y' \in \mathcal{V}} \exp(\boldsymbol{\ell}^{d}_{i,y'})}.
\end{equation}
The degraded branch is expected to assign higher probability to tokens that are plausible under weakened visual grounding, and therefore serves as a reference for identifying hallucination-prone predictions.

The contrastive logits are then computed by subtracting the degraded logits from the clean logits:
\begin{equation}
\boldsymbol{\ell}^{cd}_{i}
=
(1+\alpha)\boldsymbol{\ell}^{c}_{i}
-
\alpha \boldsymbol{\ell}^{d}_{i},
\end{equation}
where $\alpha \geq 0$ controls the contrastive strength.
Equivalently, for each candidate token $y \in \mathcal{V}$, the contrastive score is
\begin{equation}
s_i(y)
=
(1+\alpha)\boldsymbol{\ell}^{c}_{i,y}
-
\alpha \boldsymbol{\ell}^{d}_{i,y}.
\end{equation}
The next-token distribution is obtained by applying softmax to the contrastive logits:
\begin{equation}
p_{cd}(y_i=y \mid v,v_d,t,y_{<i})
=
\frac{\exp(s_i(y))}
{\sum_{y' \in \mathcal{V}} \exp(s_i(y'))}.
\end{equation}
The next token is then selected as
\begin{equation}
y_i
=
\arg\max_{y \in \mathcal{V}} p_{cd}(y \mid v,v_d,t,y_{<i}),
\end{equation}
or sampled from $p_{cd}$ depending on the decoding strategy.

In practice, contrastive decoding is often combined with a clean-branch plausibility constraint to avoid over-promoting tokens that receive very low probability under the original visual input.
Let $\beta \in [0,1]$ denote a plausibility threshold.
The candidate set can be defined as
\begin{equation}
\mathcal{V}_{i}^{\mathrm{head}}
=
\left\{
y \in \mathcal{V}
\mid
p^{c}_{i}(y)
\geq
\beta \max_{y' \in \mathcal{V}} p^{c}_{i}(y')
\right\}.
\end{equation}
The final decoding decision is then restricted to this candidate set:
\begin{equation}
y_i
=
\arg\max_{y \in \mathcal{V}_{i}^{\mathrm{head}}}
s_i(y).
\end{equation}
This constraint ensures that the contrastive objective mainly suppresses tokens over-favored by the degraded branch, rather than introducing unlikely tokens that are not supported by the clean branch.

Intuitively, the clean logits contain both visually grounded predictions and hallucination-related components, while the degraded logits are intended to emphasize predictions that remain plausible when visual grounding is weakened.
By subtracting the degraded logits, contrastive decoding suppresses tokens that rely more on priors or insufficient grounding than on reliable visual evidence.
Conventional input-level methods construct $v_d$ by perturbing the image or instruction, e.g., through image noise, masking, or instruction perturbation.
Such methods typically require an additional degraded forward pass:
\begin{equation}
\boldsymbol{\ell}^{c}_{i}=f_{\theta}(v,t,y_{<i}),
\qquad
\boldsymbol{\ell}^{d}_{i}=f_{\theta}(v_d,t,y_{<i}),
\end{equation}
which doubles a large portion of the inference computation.
In contrast, our method constructs the degraded branch inside the decoder, allowing the clean and degraded branches to share early-layer computation while still producing contrastive logits for hallucination mitigation.

\definecolor{yardblue}{RGB}{35,75,150}
\definecolor{yardorange}{RGB}{230,120,25}
\definecolor{yardgreen}{RGB}{230,244,222}
\definecolor{yardred}{RGB}{210,45,45}

\begin{figure*}[h]
\centering
\begin{tcolorbox}[
    enhanced,
    width=0.96\textwidth,
    colback=yardgreen,
    colframe=yardgreen,
    arc=18pt,
    boxrule=0pt,
    left=10pt,
    right=10pt,
    top=10pt,
    bottom=10pt
]

\begin{minipage}[t]{0.32\textwidth}
    \centering
    \textcolor{yardblue}{\textbf{\textit{Input image}}}\\[4pt]
    \includegraphics[width=\linewidth]{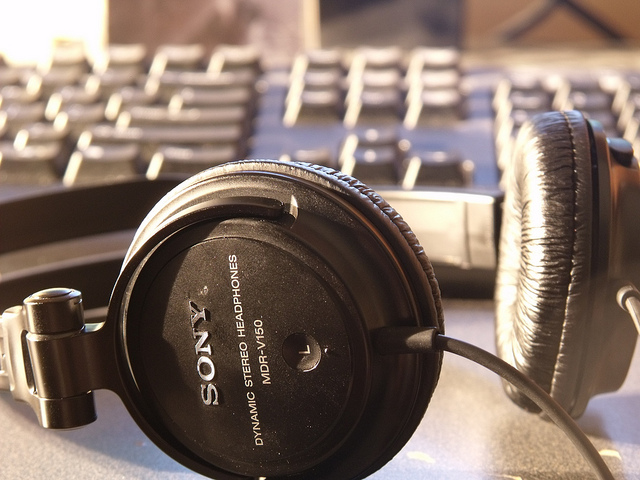}

    \vspace{6pt}
    \scriptsize
    \raggedright
    \textbf{Hallucinated object:} \hall{chair}\\
    \textbf{Prefix:} ``The image depicts a well-lit workstation with a black''\\
    \textbf{Next baseline token:} \textit{office}
\end{minipage}
\hfill
\begin{minipage}[t]{0.63\textwidth}
\scriptsize

\textcolor{yardblue}{\textbf{\textit{Top-5 next-token predictions}}}\\[-2pt]
\setlength{\tabcolsep}{4pt}
\renewcommand{\arraystretch}{1.13}
\begin{tabular}{p{0.28\linewidth}p{0.66\linewidth}}
\toprule
\textbf{Branch} & \textbf{Top-5 next tokens} \\
\midrule
Clean 
& \textit{Sony}, \textit{computer}, \textit{and}, \textit{keyboard}, \textit{pair} \\

Pixel-level deg.
& \textit{Sony}, \textit{and}, \textit{computer}, \textit{pair}, \textit{head} \\

Ours-register deg.
& \textit{computer}, \textit{and}, \textit{des}, \hall{\textit{chair}}, \textit{keyboard} \\

Text-only deg.
& \textit{des}, \textit{and}, \textit{computer}, \textit{background}, \hall{\textit{chair}} \\
\bottomrule
\end{tabular}

\vspace{8pt}

\textcolor{yardblue}{\textbf{\textit{Distributional gap to the clean branch}}}\\[-2pt]
\setlength{\tabcolsep}{5pt}
\renewcommand{\arraystretch}{1.12}
\begin{tabular}{lcc}
\toprule
\textbf{Branch vs. clean} & \textbf{Top-10 overlap} & \textbf{SymKL} \\
\midrule
Pixel-level deg. & 0.80 & 0.22 \\
Ours-register deg. & 0.60 & 7.80 \\
Text-only deg. & 0.40 & 9.20 \\
\bottomrule
\end{tabular}

\vspace{8pt}

\begin{tcolorbox}[
    enhanced,
    colback=white,
    colframe=yardblue!35,
    boxrule=0.45pt,
    arc=3pt,
    left=5pt,
    right=5pt,
    top=4pt,
    bottom=4pt
]
\scriptsize
Pixel-level degradation largely copies the clean distribution.
Text-only degradation drifts toward generic language-prior tokens such as \textit{background}.
In contrast, the register-based branch remains scene-aware by retaining tokens such as \textit{computer} and \textit{keyboard}, while exposing the hallucination-prone token \hall{\textit{chair}}.
\end{tcolorbox}

\end{minipage}

\end{tcolorbox}

\caption{
Next-token case study comparing clean and degraded branches.
Pixel-level degradation stays close to the clean visual-context distribution, while text-only degradation moves toward image-agnostic language priors.
The register-based degraded branch remains related to the visual scene while surfacing the hallucination-prone token \textit{chair}, illustrating its image-aware but locally under-grounded behavior.
}
\label{fig:top_token_case}
\end{figure*}

\subsection{Top-Token Case Study of Degraded Branches}
\label{app:top_token_case}

To provide a more intuitive view of how different degraded branches behave during decoding, we present a next-token case study in Figure~\ref{fig:top_token_case}. 
This example compares the clean branch with three degraded branches at the same decoding step, where the baseline response eventually hallucinates the object \textit{chair}. 
The prefix before prediction is ``The image depicts a well-lit workstation with a black'', and we inspect the top-5 next-token predictions from each branch.

The pixel-level degraded branch produces top predictions that are highly similar to the clean branch. 
For example, both branches assign high ranks to scene-relevant tokens such as \textit{Sony}, \textit{computer}, and \textit{keyboard}, and their Top-10 overlap remains high. 
This suggests that pixel-level degradation does not sufficiently separate the degraded branch from the clean visual-context distribution. 
As a result, its degraded logits may still contain substantial correct visual evidence, making contrastive subtraction less precise.

The text-only degraded branch shows the opposite behavior. 
It moves farther away from the clean branch and surfaces the hallucination-prone token \textit{chair}, but it also drifts toward generic language-prior completions such as \textit{background}. 
Since this branch is no longer conditioned on the image, its top predictions are less aligned with the current visual scene. 
Thus, although text-only degradation can expose hallucination-prone tokens, the resulting contrastive signal is overly image-agnostic.

In contrast, the register-based degraded branch provides a more balanced behavior. 
It still preserves image-related tokens such as \textit{computer} and \textit{keyboard}, showing that it remains semantically connected to the visual scene. 
Meanwhile, it also exposes the hallucination-prone token \textit{chair}, indicating that local visual grounding has been weakened. 
This supports our motivation for feature-level degradation: the degraded branch should be image-aware but locally under-grounded, rather than either copying the clean distribution or collapsing into a text-only prior.

\paragraph{Potential Risks.}
YARD is designed to mitigate hallucinations in LVLMs, but it does not eliminate all visually unsupported generations. In safety-critical applications, such as medical, legal, or autonomous decision-making scenarios, its outputs should still be verified by humans or external tools. In addition, because YARD modifies decoding behavior without retraining, its effectiveness may vary across model architectures, prompts, and visual domains. A potential risk is over-reliance on reduced hallucination rates as a guarantee of factual correctness. We therefore recommend using YARD as a complementary inference-time mitigation method rather than as a standalone safety mechanism.

\section{Implementation Details of Training-Free Register Construction}
\label{app:register_implementation}

\paragraph{Register construction on LLaVA-v1.5.}
In our main LLaVA-v1.5 setting, we do not train new register tokens. 
Instead, we construct them on the fly inside the CLIP ViT visual tower using forward hooks. 
During the register discovery stage, we use \texttt{register\_discovery\_max\_images=128} calibration images. 
For each image, we append \texttt{num\_register\_tokens=$M$} zero-valued register slots to the end of the original visual token sequence, and register forward hooks on the \texttt{mlp.fc1} module of each ViT layer to record the intermediate activations.

To identify register-related dimensions, we score neurons by averaging the absolute \texttt{fc1} activations at anomalous token positions across layers and calibration images. 
We then select the top \texttt{register\_topk\_neurons=10} globally activated neurons and store them in a neuron cache. 
This cache is computed once and reused during inference.

At inference time, we load the cached neuron indices and again append zero register slots to the visual token sequence. 
In the corresponding \texttt{fc1} hooks, we apply sink shift with \texttt{register\_intervention\_scale=1.0}: for the selected neuron dimensions, the strongest sink activations are redirected to the appended register slots, while ordinary patch tokens keep their original values under \texttt{register\_normal\_values=same}. 
The resulting register tokens are then passed through the multimodal projector to align them with the LLM hidden dimension.

For contrastive decoding, we use \texttt{merge\_pool\_mode=register} to construct the degraded visual condition from the projected register representations. 
The clean branch retains the full set of original visual tokens. 
The two branches share the shallow LLM decoder computation and split at \texttt{split\_layer=10}; after branching, the degraded branch is forwarded for \texttt{num\_degraded\_layers=22} layers. 
This implementation constructs the degraded branch without retraining and avoids a complete second forward pass.

\begin{figure*}[t]
\centering
\begin{tcolorbox}[
    enhanced,
    width=0.96\textwidth,
    colback=yardgreen,
    colframe=yardgreen,
    arc=18pt,
    boxrule=0pt,
    left=10pt,
    right=10pt,
    top=10pt,
    bottom=10pt
]

\begin{minipage}[c]{0.34\textwidth}
    \centering
    \textcolor{yardblue}{\textbf{\textit{Input image}}}\\[4pt]
    \includegraphics[width=\linewidth]{latex/rank02_basechairi0.750_h3_COCO_val2014_000000304817.jpg}
\end{minipage}
\hfill
\begin{minipage}[c]{0.60\textwidth}
\small

\textcolor{yardblue}{\textbf{\textit{Baseline}}}\\[-2pt]
\textit{
The image features a \hall{skateboard} next to a tree, with some leaves covering a portion of the snowboard.
The \hall{skateboard} is mostly located on top of a fence, and the tree branches reach towards it.
The fence beneath the \hall{skateboard} also has a \hall{camera} placed at the bottom of the image.
}

\vspace{12pt}

\textcolor{yardblue}{\textbf{\textit{YARD}}}\\[-2pt]
\textit{
The image features a snowboard leaning against a fence with green leaves in the background.
The snowboard is quite large and is resting on a wooden fence.
It has a blue design and is placed on its side, making it stand out against the green foliage.
The fence is visible in the background, partially covered by leaves, while the snowboard remains the main object in the scene.
}

\end{minipage}

\end{tcolorbox}

\caption{
Qualitative comparison between the baseline and YARD.
The baseline repeatedly hallucinates \textit{skateboard} and introduces an unsupported \textit{camera}, while YARD correctly describes the visible snowboard and avoids unsupported objects.
Hallucinated words are highlighted in red.
}
\label{fig:snowboard_case}
\end{figure*}

\begin{figure*}[t]
\centering
\begin{tcolorbox}[
    enhanced,
    width=0.96\textwidth,
    colback=yardgreen,
    colframe=yardgreen,
    arc=18pt,
    boxrule=0pt,
    left=10pt,
    right=10pt,
    top=10pt,
    bottom=10pt
]

\begin{minipage}[c]{0.34\textwidth}
    \centering
    \textcolor{yardblue}{\textbf{\textit{Input image}}}\\[4pt]
    \includegraphics[width=\linewidth]{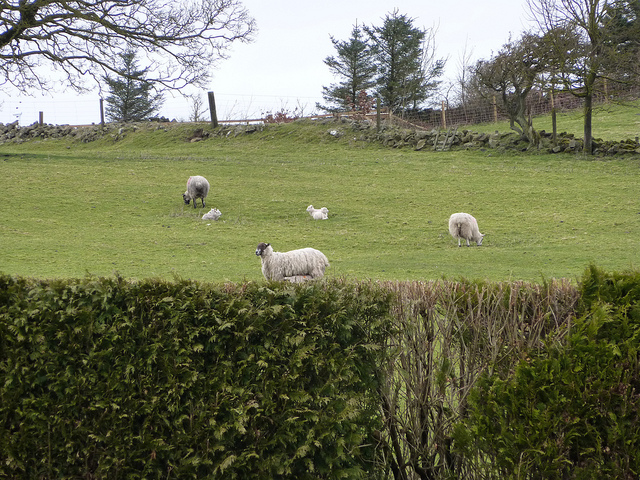}
\end{minipage}
\hfill
\begin{minipage}[c]{0.60\textwidth}
\small

\textcolor{yardblue}{\textbf{\textit{Baseline}}}\\[-2pt]
\textit{
The scene depicts a large, peaceful, grassy field where a mixed herd of sheep and baby lambs is grazing.
There are \hall{people} observing the flock from a nearby bush or landscaped position.
The sheep are scattered throughout the field, with some lambs huddled together while others graze farther across the hillside.
}

\vspace{12pt}

\textcolor{yardblue}{\textbf{\textit{YARD}}}\\[-2pt]
\textit{
The image depicts a peaceful scene of several sheep grazing on a lush green hillside.
There are ten sheep in total, with some standing and others lying down on the grassy field.
The sheep are scattered throughout the field, some closer to the foreground, while others are further in the background.
The overall scene is calm and natural, with the green hillside providing a simple backdrop for the flock.
}

\end{minipage}

\end{tcolorbox}

\caption{
Qualitative comparison between the baseline and YARD.
The baseline hallucinates unsupported \textit{people}, while YARD focuses on visually grounded content and correctly describes the flock of sheep on the grassy hillside.
Hallucinated words are highlighted in red.
}
\label{fig:sheep_case}
\end{figure*}

\begin{figure*}[t]
\centering
\begin{tcolorbox}[
    enhanced,
    width=0.96\textwidth,
    colback=yardgreen,
    colframe=yardgreen,
    arc=18pt,
    boxrule=0pt,
    left=10pt,
    right=10pt,
    top=10pt,
    bottom=10pt
]

\begin{minipage}[c]{0.34\textwidth}
    \centering
    \textcolor{yardblue}{\textbf{\textit{Input image}}}\\[4pt]
    \includegraphics[width=\linewidth]{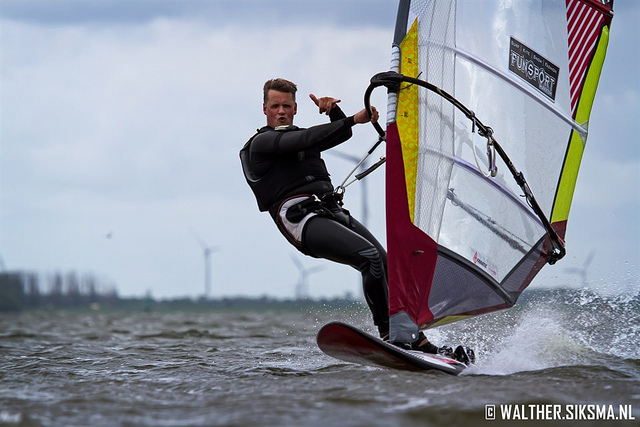}
\end{minipage}
\hfill
\begin{minipage}[c]{0.60\textwidth}
\small

\textcolor{yardblue}{\textbf{\textit{Baseline}}}\\[-2pt]
\textit{
The image features a man skillfully riding a board in ocean waves while holding onto a sail, demonstrating his expertise in kite surfing.
He is wearing a wetsuit and is completely engaged in the thrilling watersport.
The \hall{kite} is visible in the wind's motion, providing power for the rider as he moves quickly across the water.
}

\vspace{12pt}

\textcolor{yardblue}{\textbf{\textit{YARD}}}\\[-2pt]
\textit{
The image captures a man skillfully windsurfing on a body of water.
He is wearing a black wet suit and riding a red and black wind surfboard.
The windsurfer is leaning back as he enjoys the thrill of the sport, with the sail attached to the board clearly visible and helping propel him across the water.
}

\end{minipage}

\end{tcolorbox}

\caption{
Qualitative comparison between the baseline and YARD.
The baseline hallucinates an unsupported \textit{kite}, while YARD correctly focuses on the visually grounded windsurfing scene, including the person, the sail, and the surfboard.
Hallucinated words are highlighted in red.
}
\label{fig:windsurf_case}
\end{figure*}

\begin{figure*}[t]
\centering
\begin{tcolorbox}[
    enhanced,
    width=0.96\textwidth,
    colback=yardgreen,
    colframe=yardgreen,
    arc=18pt,
    boxrule=0pt,
    left=10pt,
    right=10pt,
    top=10pt,
    bottom=10pt
]

\begin{minipage}[c]{0.34\textwidth}
    \centering
    \textcolor{yardblue}{\textbf{\textit{Input image}}}\\[4pt]
    \includegraphics[width=\linewidth]{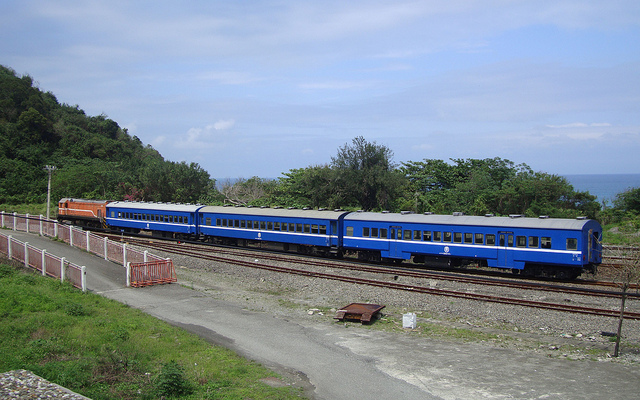}
\end{minipage}
\hfill
\begin{minipage}[c]{0.60\textwidth}
\small

\textcolor{yardblue}{\textbf{\textit{Baseline}}}\\[-2pt]
\textit{
The image features a bright blue passenger train traveling near a forest.
It occupies almost the entire visible width of the scene, running parallel to the tracks near the road sign.
There are six visible \hall{cars} on the train, each with a clear blue color and easy to distinguish by their alignment.
The train is moving steadily along the railway, with the surrounding greenery providing a natural backdrop.
}

\vspace{12pt}

\textcolor{yardblue}{\textbf{\textit{YARD}}}\\[-2pt]
\textit{
The image features a long blue passenger train traveling on the tracks.
The train is stretched out across the scene, covering a significant portion of the image.
The tracks appear to be made of metal, and the train is moving along them in a straight line.
In addition to the train, there is surrounding greenery and a signpost visible near the railway, completing the outdoor scene.
}

\end{minipage}

\end{tcolorbox}

\caption{
Qualitative comparison between the baseline and YARD.
The baseline hallucinates unsupported \textit{cars} as separate objects, while YARD consistently focuses on the visually grounded \textit{train} and its surrounding scene.
Hallucinated words are highlighted in red.
}
\label{fig:train_case}
\end{figure*}

\begin{figure*}[t]
\centering
\begin{tcolorbox}[
    enhanced,
    width=0.96\textwidth,
    colback=yardgreen,
    colframe=yardgreen,
    arc=18pt,
    boxrule=0pt,
    left=10pt,
    right=10pt,
    top=10pt,
    bottom=10pt
]

\begin{minipage}[c]{0.34\textwidth}
    \centering
    \textcolor{yardblue}{\textbf{\textit{Input image}}}\\[4pt]
    \includegraphics[width=\linewidth]{images/rank09_basechairi0.500_h2_COCO_val2014_000000061108.jpg}
\end{minipage}
\hfill
\begin{minipage}[c]{0.60\textwidth}
\small

\textcolor{yardblue}{\textbf{\textit{Baseline}}}\\[-2pt]
\textit{
The scene displays a small dog sitting inside a wire basket placed on a transport \hall{motorcycle}.
The dog appears to be comfortable in its enclosed space, offering a sense of security.
Behind the \hall{motorcycle}, there is a trash can with a cardboard box on it.
Near the edge of the image, additional street-side objects can also be seen.
}

\vspace{12pt}

\textcolor{yardblue}{\textbf{\textit{YARD}}}\\[-2pt]
\textit{
The scene features a dog sitting inside a wire basket attached to the front of a blue bicycle.
The dog appears to be a black and white terrier, comfortably occupying the basket.
The bicycle is parked next to a garbage can, which is located on the right side of the image.
A bench and a car can also be seen in the surrounding street scene.
}

\end{minipage}

\end{tcolorbox}

\caption{
Qualitative comparison between the baseline and YARD.
The baseline hallucinates an unsupported \textit{motorcycle}, while YARD correctly identifies the \textit{bicycle} and preserves other visually grounded content such as the dog, the garbage can, and the surrounding street scene.
Hallucinated words are highlighted in red.
}
\label{fig:dog_bicycle_case}
\end{figure*}

\begin{figure*}[t]
\centering
\begin{tcolorbox}[
    enhanced,
    width=0.96\textwidth,
    colback=yardgreen,
    colframe=yardgreen,
    arc=18pt,
    boxrule=0pt,
    left=10pt,
    right=10pt,
    top=10pt,
    bottom=10pt
]

\begin{minipage}[c]{0.34\textwidth}
    \centering
    \textcolor{yardblue}{\textbf{\textit{Input image}}}\\[4pt]
    \includegraphics[width=\linewidth]{images/rank15_basechairi0.400_h2_COCO_val2014_000000368349.jpg}
\end{minipage}
\hfill
\begin{minipage}[c]{0.60\textwidth}
\small

\textcolor{yardblue}{\textbf{\textit{Baseline}}}\\[-2pt]
\textit{
The image shows a clean bathroom that features a white toilet under a towel rail.
Toilet paper can be seen on the wall, and there is a \hall{toothbrush} holder with \hall{toothbrushes}.
Above the toilet, a pull-down window blind helps control the light entering the room, while other bathroom items are arranged around the sink area.
}

\vspace{12pt}

\textcolor{yardblue}{\textbf{\textit{YARD}}}\\[-2pt]
\textit{
The image shows a bathroom with white towels hanging from a towel rack on the wall.
The towels are folded neatly, and there are four of them, with two on each side.
The toilet is located in the middle of the bathroom, with a window nearby and a sink area visible in the surrounding space.
The scene appears clean and orderly, with simple bathroom fixtures arranged along the wall.
}

\end{minipage}

\end{tcolorbox}

\caption{
Qualitative comparison between the baseline and YARD.
The baseline hallucinates unsupported \textit{toothbrushes}, while YARD focuses on visually grounded bathroom elements such as the toilet, towels, window, and sink area.
Hallucinated words are highlighted in red.
}
\label{fig:bathroom_case}
\end{figure*}

\end{document}